\documentclass[11pt]{article}

\usepackage[final]{colm2026_conference}

\usepackage[T1]{fontenc}
\usepackage[table]{xcolor}
\definecolor{gold}{HTML}{FFD700}
\definecolor{silver}{HTML}{C0C0C0}
\usepackage[utf8]{inputenc}
\usepackage{microtype}
\usepackage{graphicx}
\usepackage{longtable}
\usepackage{siunitx}
\usepackage{array}
\usepackage{ragged2e}
\usepackage{booktabs}
\usepackage{pdflscape}
\usepackage{amsmath}
\usepackage{amssymb}
\usepackage{mathtools}
\usepackage{hyperref}
\usepackage{float}
\usepackage[capitalize,noabbrev]{cleveref}
\newcommand{\gbest}[1]{\cellcolor{gold!30}\textbf{#1}}
\newcommand{\sbest}[1]{\cellcolor{silver!40}#1}

\title{Multilingual Prompt Localization for Agent-as-a-Judge: Language and Backbone Sensitivity in Requirement-Level Evaluation}
\author{
Alhasan Mahmood$^{1,*}$, Samir Abdaljalil$^{2,*}$, Hasan Kurban$^{1}$ \\
\\
$^{1}$ Hamad Bin Khalifa University, Doha, Qatar \\
$^{2}$ Texas A\&M University, College Station, TX USA \\
$^{*}$ Equal contribution \\
\texttt{Corresponding Author: hkurban@hbku.edu.qa}
}

\begin{document}
\maketitle

\begin{abstract}
Evaluation language is typically treated as a fixed English default in agentic code benchmarks, yet we show that changing the judge's language can invert backbone rankings. We localize the Agent-as-a-Judge prompt stack to five typologically diverse languages (English, Arabic, Turkish, Chinese, Hindi) and evaluate 55 DevAI development tasks across three developer-agent frameworks and six judge backbones, totaling 4950 judge runs. The central finding is that backbone and language interact: GPT-4o achieves the highest satisfaction in English (44.72\%), while Gemini leads in Arabic (51.72\%, $p<0.001$ vs.\ GPT-4o) and Hindi (53.22\%). No single backbone dominates across all languages, and inter-backbone agreement on individual requirement judgments is modest (Fleiss' $\kappa \leq 0.231$). A controlled ablation further shows that localizing judge-side instructions, not just benchmark content, can be decisive: Hindi satisfaction drops from 42.8\% to 23.2\% under partial localization. These results indicate that language should be treated as an explicit evaluation variable in agentic benchmarks. Full requirement-level judgments and runtime statistics are released for reproducibility.
\end{abstract}

\section{Introduction}

Evaluation of coding agents relies heavily on final-outcome benchmarks, despite the process-driven nature of long-horizon development tasks. Agent-as-a-Judge (AAAJ) addresses this by enabling evaluators to inspect intermediate artifacts rather than relying only on final outputs \citep{zhuge2024agentjudge}, but its judging stack is English-centric, which risks evaluation distortion in multilingual deployment.

We document a previously unmeasured confound in agentic evaluation: evaluation language and judge backbone interact in ways that change backbone rankings. Just as annotation artifacts \citep{gururangan2018annotation} and benchmark contamination \citep{magar2022data} revealed hidden confounds that altered community practices for NLU evaluation, we show that the language of the judge prompt is a confound that can invert the relative ordering of backbone models. This is a measurement contribution: we do not propose a new evaluation method, but provide systematic evidence that an implicit assumption of current benchmarks (English-only judging) introduces a measurable and consequential bias.

Concretely, we localize the AAAJ prompt stack to five typologically diverse languages (English, Arabic, Turkish, Chinese, and Hindi) and evaluate 55 DevAI tasks across three developer-agent frameworks (MetaGPT \citep{hong2024metagpt}, GPT-Pilot \citep{pythagora2023gptpilot}, OpenHands \citep{wang2024opendevin}) and six judge backbones (GPT-4o, GPT-5.4, Claude Sonnet 4.6, Gemini 3 Flash Preview, DeepSeek V3.2, Qwen3.5-9B), totaling 4950 judge runs.

\begin{figure*}[t]
\centering
\includegraphics[width=\linewidth]{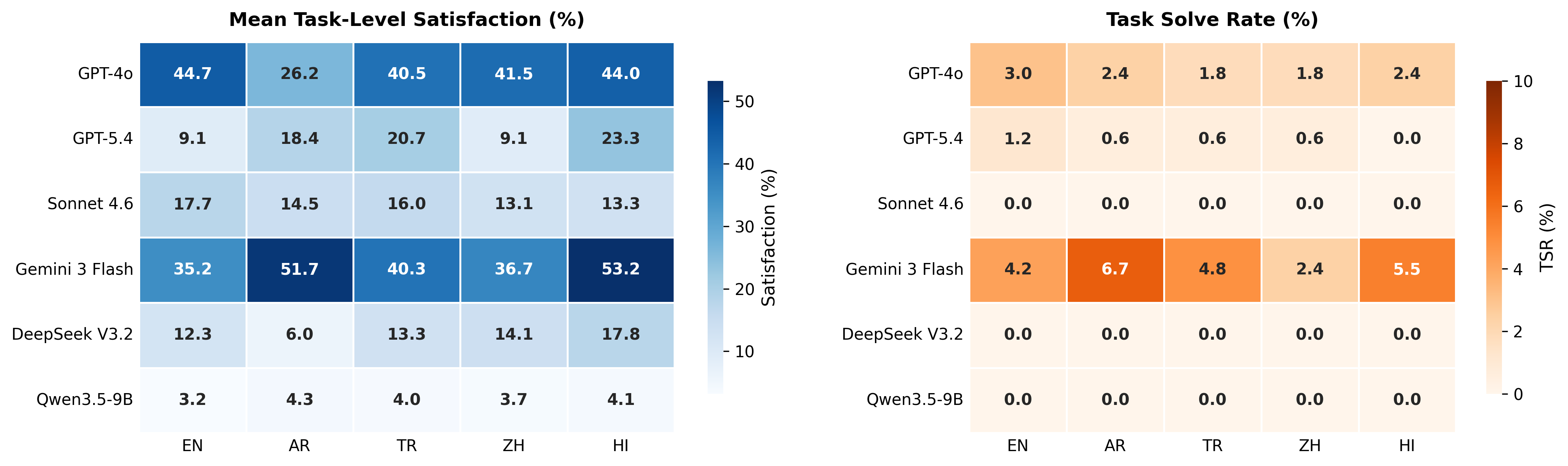}
\caption{Backbone-by-language heatmaps averaged over three developer-agent frameworks. \textbf{Left:} Mean task-level requirement satisfaction (\%), using a sequential blue colormap where darker shading indicates higher satisfaction. \textbf{Right:} Task Solve Rate (\%), using a sequential orange colormap where darker shading indicates higher completion. Each cell displays the numeric value. The strongest backbone varies across languages: GPT-4o leads in English (44.7\%), while Gemini leads in Arabic (51.7\%) and Hindi (53.2\%), showing that evaluation outcomes depend jointly on language and backbone.}
\label{fig:intro_backbone_language_heatmap}
\end{figure*}

Figure~\ref{fig:intro_backbone_language_heatmap} previews the central finding: backbone rankings change across languages, meaning that English-only evaluation can produce misleading comparisons for multilingual deployment.

Our contributions are as follows.
(1) We provide the first systematic evidence that evaluation language and judge backbone interact in ways that change backbone rankings, with GPT-4o strongest in English but Gemini strongest in Arabic ($p<0.001$) and Hindi.
(2) We present a multilingual benchmark study over 55 development tasks and 365 requirement checks per language-framework pair across six backbones, totaling 4950 judge runs.
(3) We introduce a practical method for localizing the AAAJ prompt stack while preserving its structure and evidence-grounded workflow, and show through ablation that instruction-language localization can be decisive for evaluation stability.
(4) We release summary metrics, full requirement-level judgments, and task-level runtime statistics for transparency and reproducibility, following the auditable design of \citet{zhuge2024agentjudge}.

\section{Related Work}

\textbf{AI judges and agentic evaluation.} Our work builds on Agent-as-a-Judge, which extends LLM-as-a-Judge to inspect intermediate artifacts rather than only final outputs \citep{zhuge2024agentjudge}. Prior work demonstrates the usefulness of model-based evaluation but also highlights bias, instability, and limited faithfulness under output-only judging \citep{zheng2024judging}. Fine-grained approaches such as FLASK improve interpretability via skill decomposition \citep{ye2024flask}. Multilingual evaluation has been studied through cross-lingual meta-evaluation \citep{doddapaneni2025cross} and evidence of cross-language inconsistency in English-strong evaluators \citep{son2024mmeval,fu2025reliable}, with further work on language-agnostic criteria transfer \citep{sheth2026crosslingual}. Alternative paradigms such as PAJAMA recast evaluation as program synthesis \citep{huang2025pajama}. Unlike CIA/Hercule, which trains a dedicated multilingual evaluator, we study how existing monolingual backbones respond to prompt localization without retraining, revealing backbone-language interactions not captured by prior work.

\textbf{Benchmarks for code and ML agents.} Early benchmarks such as HumanEval \citep{chen2021codex} and MBPP \citep{austin2021program} focus on short program synthesis, while SWE-Bench \citep{jimenez2023swebench} and MLE-Bench \citep{chan2024mlebench} move toward realistic workflows. Recent benchmarks expand to repository-grounded tasks \citep{ni2025gittaskbench}, broader execution settings \citep{xu2025swecompass}, and diverse engineering activities \citep{sonwane2026omnicode}. DevAI differs by structuring tasks as hierarchical requirements, enabling requirement-level analysis \citep{zhuge2024agentjudge}. Unlike DevAI's original single-backbone English evaluation, we vary both language and backbone to expose interaction effects that single-configuration evaluations cannot detect.

\textbf{Developer-agent frameworks and multilinguality.} We study representative developer-agent paradigms: multi-agent decomposition (MetaGPT) \citep{hong2024metagpt}, tool-oriented workflows (GPT-Pilot) \citep{pythagora2023gptpilot}, and interactive agents (OpenHands) \citep{wang2024opendevin}. Multilingual code generation has been explored in MultiPL-E \citep{cassano2023multipl} and extended in mHumanEval \citep{raihan2025mhumaneval} and McEval \citep{chai2025mceval}. Unlike these benchmarks, which study multilingual code \textit{generation}, we study multilingual \textit{evaluation} of code agents on long-horizon development tasks, a setting where judge behavior introduces an additional source of cross-language variation.

\section{A Multilingual DevAI Benchmark}

\begin{figure*}[t]
\centering
\includegraphics[width=\linewidth]{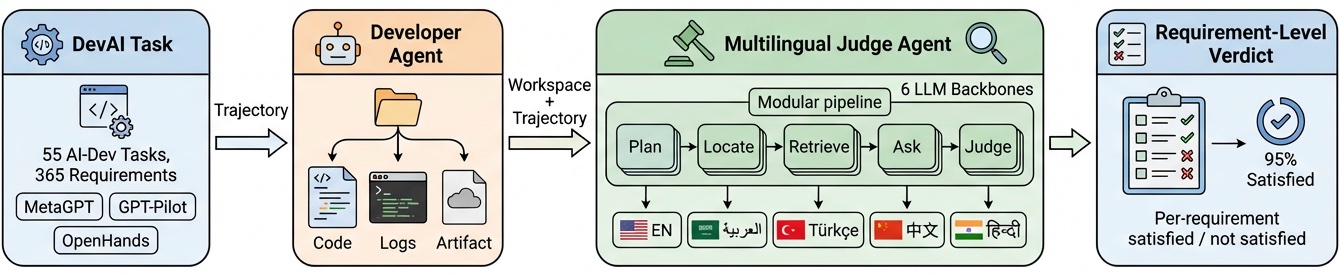}
\caption{Overview of the multilingual Agent-as-a-Judge pipeline. DevAI tasks are executed by three developer-agent frameworks whose workspaces, code, and trajectories are passed to a multilingual judge agent. The judge runs a modular pipeline (Plan $\to$ Locate $\to$ Retrieve $\to$ Ask $\to$ Judge) in each of five languages under six LLM backbones, producing per-requirement satisfaction verdicts.}
\label{fig:pipeline_overview}
\end{figure*}

\paragraph{Motivation.}

Benchmarks such as HumanEval \citep{chen2021codex} and MBPP \citep{austin2021program} target function-level synthesis but do not capture the broader lifecycle of realistic AI development. Subsequent benchmarks, including SWE-Bench \citep{jimenez2023swebench} and MLE-Bench \citep{chan2024mlebench}, move closer to practical software engineering, yet scalable evaluation of multi-step development remains challenging. DevAI and Agent-as-a-Judge address this by exposing intermediate requirement satisfaction rather than only final outcomes.

In multilingual settings, benchmark structure becomes even more critical. A multilingual judge must not only process translated text but also preserve development milestones, requirement dependencies, and evidence-grounded reasoning. DevAI is well suited for this purpose: tasks are decomposed into explicit, hierarchical requirements that can be evaluated independently. This enables a more precise question than ``Did the system solve the task?'', specifically, which components remain stable under localization and which degrade across languages. This aligns with the AAAJ motivation for intermediate, process-aware evaluation \citep{zhuge2024agentjudge}.

\paragraph{The Multilingual DevAI Dataset.}

We use the DevAI benchmark introduced in \citep{zhuge2024agentjudge}, consisting of 55 realistic AI development tasks spanning data processing, modeling, training, evaluation, reporting, visualization, and interaction. The benchmark includes 365 explicit requirements organized as milestone structures, often with prerequisite dependencies, enabling analysis of partial progress rather than only end-to-end success.

Our multilingual extension preserves this task topology and instead localizes the judging interface and evidence interpretation layers. Requirement text, evaluation criteria, and judge justifications are expressed in the target language, while the underlying task structure remains unchanged. Concretely, we localize the judge, ask, locate, planning, and retrieve modules. This maintains comparability with the English benchmark while isolating the effect of language on requirement-level evaluation.

We evaluate five languages chosen for typological diversity \citep{fitzgerald2023massive}: English (baseline), Arabic (right-to-left, morphologically rich \citep{habash2010arabic}), Turkish (agglutinative, Latin script \citep{ozturel2019turkish}), Chinese (non-segmented writing system \citep{huang2006chinese}), and Hindi (Devanagari script \citep{singh2024indicgenbench}). These languages stress different aspects of prompt processing (tokenization, morphology, script handling, terminology transfer) and provide a stronger robustness test than typologically closer languages. Because tasks remain fixed, observed differences isolate prompt localization effects.

\paragraph{Preliminaries and Notation.}

Let $\mathcal{T} = \{t_1, \ldots, t_{55}\}$ denote the set of DevAI tasks, where each task $t$ has a requirement set $R(t) = \{r_1, \ldots, r_{|R(t)|}\}$ with $\sum_{t} |R(t)| = 365$. Let $\mathcal{L} = \{\text{EN}, \text{AR}, \text{TR}, \text{ZH}, \text{HI}\}$ be the set of evaluation languages, $\mathcal{B}$ the set of six judge backbones, and $\mathcal{F} = \{\text{MetaGPT}, \text{GPT-Pilot}, \text{OpenHands}\}$ the set of developer-agent frameworks. For a given backbone $b \in \mathcal{B}$, language $\ell \in \mathcal{L}$, and framework $f \in \mathcal{F}$, the judge produces a binary verdict $J(t, r, \ell, b, f) \in \{0, 1\}$ for each requirement $r \in R(t)$, where $1$ denotes \texttt{satisfied}. We define:

\textit{Requirement Satisfaction Rate} for a task $t$:
\[
\text{Sat}(t, \ell, b, f) = \frac{\sum_{r \in R(t)} J(t, r, \ell, b, f)}{|R(t)|} \times 100.
\]

\textit{Task Solve Rate} (TSR): the fraction of tasks achieving $\text{Sat}(t, \ell, b, f) = 100\%$. We additionally report percentile success rates at the 20\%, 50\%, and 70\% thresholds.

\paragraph{Evaluation Protocol.}

We evaluate the DevAI benchmark across three developer-agent frameworks: \texttt{MetaGPT}, \texttt{GPT-Pilot}, and \texttt{OpenHands}. The multilingual setup covers five languages as illustrated in Figure~\ref{fig:pipeline_overview}.

The judge is instantiated with six backbone LLMs: \texttt{gpt-4o-2024-08-06}, \texttt{gpt-5.4}, \texttt{claude-sonnet-4.6}, \texttt{gemini-3-flash-preview}, \texttt{deepseek-v3.2}, and \texttt{qwen3.5-9b}\footnote{All backbones accessed via the OpenRouter API (\url{https://openrouter.ai}) using the corresponding model identifiers.}, with GPT-4o serving as the baseline.

Multilingual localization is applied to the judge-side prompt stack by translating requirement descriptions, evaluation criteria, and justification prompts from English into the four non-English languages while preserving task structure and evaluation logic. Translations are reviewed by native speakers with graduate-level training to ensure fluency and technical fidelity.

All experiments use a unified inference pipeline across backbones. Models are accessed via the OpenRouter API, with the local system responsible for orchestration and artifact collection. Inference uses temperature $= 0.0$, top-$p = 0.9$, a maximum of 2048 tokens, a 30-second timeout, and up to three retries. Prompt structure is identical across backbones, consisting of a system message and a task-specific user message.

Each backbone evaluates 825 task runs ($55 \times 5 \times 3$), for a total of 4950 runs across all six backbones.

\section{Experiments}


We evaluate all six judge backbones under the protocol described in Section~3. For each language-framework pair, we report $\text{Sat}(t, \ell, b, f)$ averaged over the 55 tasks with standard deviation ($\mu \pm \sigma$), complemented by percentile success rates at the 20\%, 50\%, 70\%, and 100\% thresholds. Statistical comparisons use two-sided Wilcoxon signed-rank tests on paired task-level scores (averaged across frameworks), with Holm-Bonferroni correction and 95\% bootstrap confidence intervals (Appendix~\ref{app:significance_tests}).

\begin{table*}[t]
\centering
\footnotesize
\renewcommand{\arraystretch}{1.3}
\setlength{\tabcolsep}{4.5pt}
\caption{Mean task-level requirement satisfaction (\%) $\pm$ standard deviation by language and backbone, pooled across three frameworks (165 runs per cell).
\colorbox{gold!30}{\strut Gold} = best in row; \colorbox{silver!40}{\strut Silver} = second best.}
\label{tab:language_backbone_satisfaction}
\begin{tabular}{@{}l rrrrrr@{}}
\toprule
\textbf{Lang.} & \textbf{GPT-4o} & \textbf{GPT-5.4} & \textbf{Sonnet 4.6} & \textbf{Gemini 3 Flash} & \textbf{DeepSeek V3.2} & \textbf{Qwen3.5-9B} \\
\midrule
EN & \gbest{44.7 {\scriptsize$\pm$21.1}} & 9.1 {\scriptsize$\pm$15.7}  & 17.7 {\scriptsize$\pm$16.1} & \sbest{35.2 {\scriptsize$\pm$24.1}} & 12.3 {\scriptsize$\pm$16.8} & 3.2 {\scriptsize$\pm$10.2} \\
AR & \sbest{26.2 {\scriptsize$\pm$19.6}} & 18.4 {\scriptsize$\pm$19.1} & 14.5 {\scriptsize$\pm$16.5} & \gbest{51.7 {\scriptsize$\pm$24.1}} & 6.0 {\scriptsize$\pm$13.0}  & 4.3 {\scriptsize$\pm$12.2} \\
TR & \gbest{40.5 {\scriptsize$\pm$20.5}} & 20.7 {\scriptsize$\pm$21.8} & 16.0 {\scriptsize$\pm$16.8} & \sbest{40.3 {\scriptsize$\pm$23.5}} & 13.3 {\scriptsize$\pm$19.2} & 4.0 {\scriptsize$\pm$11.8} \\
ZH & \gbest{41.5 {\scriptsize$\pm$22.3}} & 9.1 {\scriptsize$\pm$16.0}  & 13.1 {\scriptsize$\pm$16.4} & \sbest{36.7 {\scriptsize$\pm$23.4}} & 14.1 {\scriptsize$\pm$18.4} & 3.7 {\scriptsize$\pm$10.7} \\
HI & \sbest{44.0 {\scriptsize$\pm$19.4}} & 23.3 {\scriptsize$\pm$17.8} & 13.3 {\scriptsize$\pm$17.3} & \gbest{53.2 {\scriptsize$\pm$25.1}} & 17.8 {\scriptsize$\pm$18.4} & 4.1 {\scriptsize$\pm$12.1} \\
\bottomrule
\end{tabular}
\end{table*}

\begin{figure*}[t]
\centering
\includegraphics[width=\linewidth]{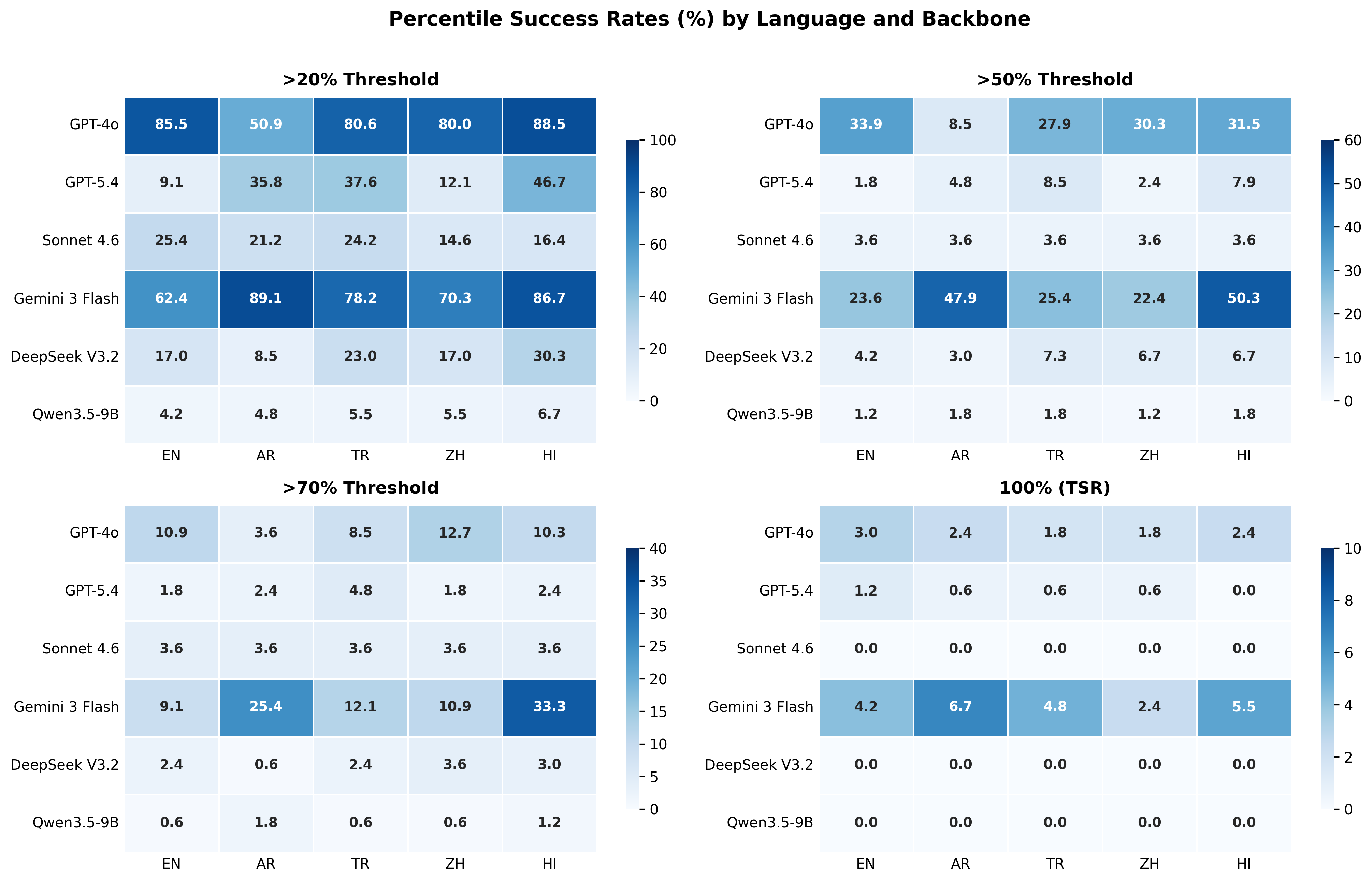}
\caption{Percentile success rates (\%) by language and backbone, averaged over three developer-agent frameworks. Each panel uses a sequential blue colormap (darker = higher fraction of tasks clearing the threshold), except the 100\% panel which uses an orange colormap for Task Solve Rate. The four panels show the fraction of tasks exceeding 20\%, 50\%, 70\%, and 100\% requirement satisfaction. GPT-4o and Gemini are similar at the 20\% threshold, but Gemini separates at harder thresholds, especially in Arabic and Hindi. The full numerical table is provided in Appendix~\ref{app:dense_tables}.}
\label{fig:percentile_success_main}
\end{figure*}

\paragraph{Language-Level Backbone Comparison.}
\textit{Does the best backbone depend on evaluation language?}

Table~\ref{tab:language_backbone_satisfaction} and Figure~\ref{fig:percentile_success_main} summarize backbone behavior aggregated across frameworks. Two patterns emerge.
First, backbone ranking depends on evaluation language. GPT-4o is strongest in English (Wilcoxon $p<0.001$, CI $[-19.05,-4.76]$ vs.\ Gemini), while Gemini achieves the highest scores in Arabic ($p<0.001$, CI $[19.05,28.57]$) and the highest mean in Hindi (53.22\%). The Chinese GPT-4o advantage is not significant ($p=0.063$), and the Gemini vs.\ GPT-4o gap in Hindi does not survive Holm-Bonferroni correction.

Second, backbone capability gates whether multilingual effects are observable: weaker backbones show near-zero cross-language variation because their baseline performance leaves little room for language to matter. Large standard deviations across all settings indicate substantial task-level heterogeneity.

The percentile view refines this picture. GPT-4o and Gemini are similar at low thresholds, but Gemini separates at higher thresholds, particularly in Arabic and Hindi (exceeding 50\% satisfaction on 50.30\% of Hindi tasks and 70\% on 33.33\%). The remaining backbones are concentrated at lower thresholds, as detailed in the per-backbone analysis below.

\paragraph{Requirement-Type Sensitivity.}
\textit{Are some requirement types more fragile under multilingual evaluation?}

\begin{figure*}[t]
\centering
\includegraphics[width=\linewidth]{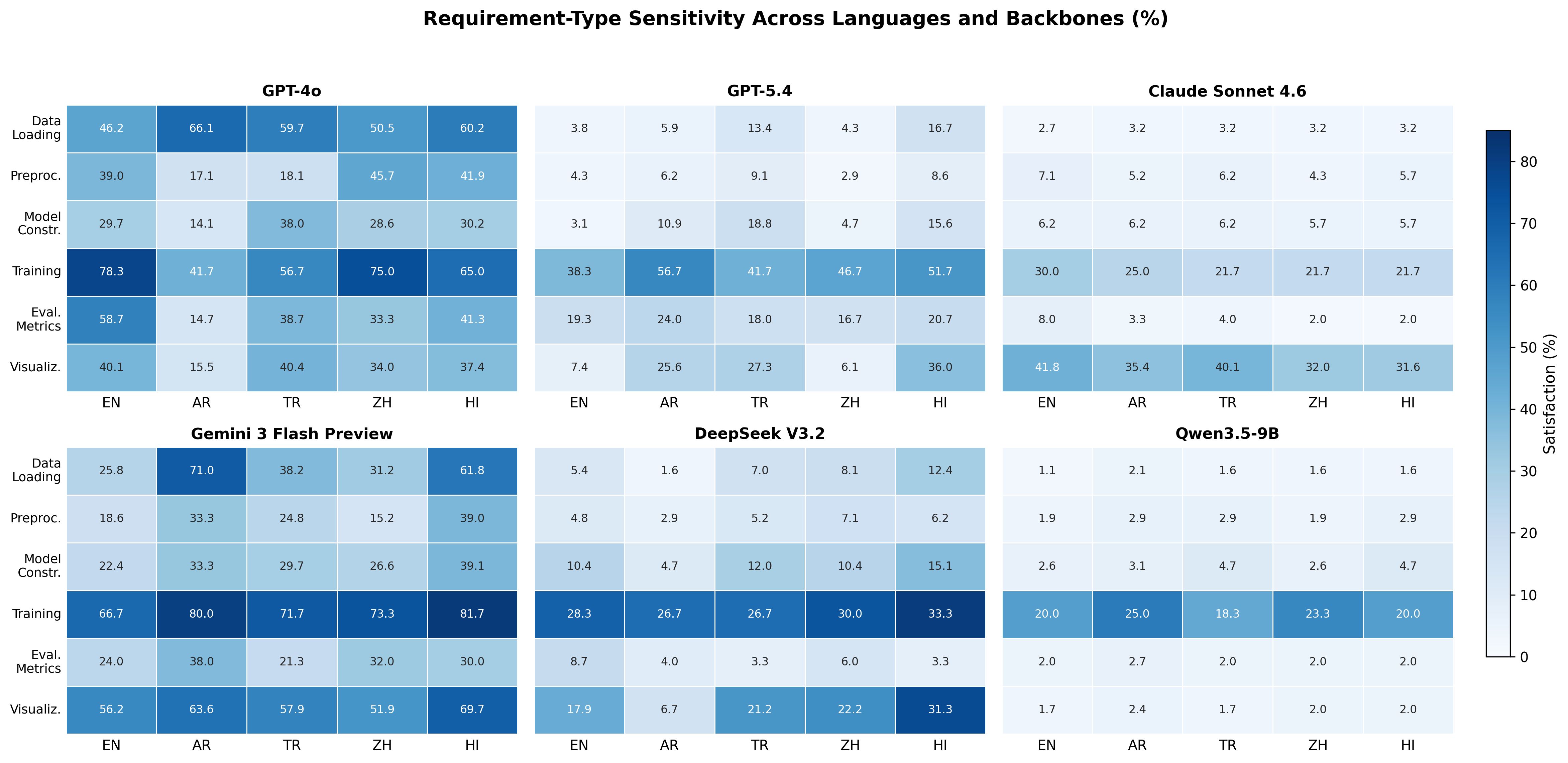}
\caption{Requirement-type sensitivity (\%) across languages and backbones. Each of the six panels shows one backbone as a heatmap, with rows corresponding to six requirement categories and columns to five evaluation languages. A sequential blue colormap encodes satisfaction rate: darker cells indicate higher satisfaction within that requirement type and language. \textit{Training} and \textit{Data Loading} are consistently the easiest categories, while \textit{Evaluation Metrics} and \textit{Model Construction} show the largest cross-language variation under strong backbones. Weaker backbones (DeepSeek, Qwen) remain near floor across all categories. The full numerical table is provided in Appendix~\ref{app:dense_tables}.}
\label{fig:requirement_type_sensitivity}
\end{figure*}

Figure~\ref{fig:requirement_type_sensitivity} reveals systematic differences across requirement types. Operational categories (\textit{Data Loading}, \textit{Training}) are consistently easier, while semantically dense categories (\textit{Model Construction}, \textit{Evaluation Metrics}) are more sensitive to language.
Under GPT-4o, \textit{Evaluation Metrics} shows the largest cross-language variation (14.67\% in Arabic vs.\ 58.67\% in English), followed by \textit{Model Construction}. In contrast, \textit{Data Loading} remains relatively stable. When weaker backbones are included, variability increases even for operational categories, as low-capability judges fail to consistently interpret straightforward requirements.

\paragraph{Model-Specific Results.}
\textit{How do individual backbones respond to multilingual evaluation?}

The six backbones separate into three performance tiers, each illustrating a different aspect of multilingual judge behavior.

\textit{Top tier: GPT-4o and Gemini.}
GPT-4o is the strongest baseline but highly variable, ranging from approximately 25\% (Arabic) to 46\% (other languages). Despite broad partial progress, language-level TSR remains low, ranging from 1.82\% to 3.03\% across the five languages (Table~\ref{tab:language_backbone_success}). Gemini achieves the highest average satisfaction, exceeding 50\% in Arabic and Hindi, with language-level TSR up to 6.67\% (Arabic; Table~\ref{tab:language_backbone_success}). The central finding is that their rankings invert across languages: GPT-4o leads in English while Gemini leads in Arabic and Hindi.

\textit{Mid tier: GPT-5.4 and Claude Sonnet 4.6.}
GPT-5.4 performs well below GPT-4o (8\% to 10\% in English and Chinese). Claude exhibits low but uniform performance (12\% to 18\%), with TSR at 0.00\% throughout, reflecting a conservative evaluation policy that applies consistently regardless of language.

\textit{Floor tier: DeepSeek V3.2 and Qwen3.5-9B.}
DeepSeek (6.02\% to 17.77\%) and Qwen (3.20\% to 4.32\%) never achieve full task completion. Their minimal cross-language variation confirms that multilingual sensitivity is only observable when baseline capability is sufficient.

\begin{table*}[t]
\centering
\footnotesize
\renewcommand{\arraystretch}{1.3}
\setlength{\tabcolsep}{3.8pt}
\caption{Aggregate backbone comparison (825 runs each). Satisfaction and cost report mean $\pm$ s.d.; success columns report \% of tasks above each threshold. \colorbox{gold!30}{\strut Gold} = best; \colorbox{silver!40}{\strut Silver} = second best. For time and tokens, $\downarrow$ = lower is better.}
\label{tab:backbone_overall}
\begin{tabular}{@{}l r rrrr rrr@{}}
\toprule
 & \textbf{Sat.\,(\%)} & \multicolumn{4}{c}{\textbf{Success Thresholds\,(\%)}} & \textbf{Time\,(s)\,$\downarrow$} & \textbf{In\,Tok.\,$\downarrow$} & \textbf{Out\,Tok.\,$\downarrow$} \\
\cmidrule(lr){3-6}
 & & {$>$20} & {$>$50} & {$>$70} & {100} & & & \\
\midrule
GPT-4o       & \sbest{39.4 {\scriptsize$\pm$21.7}} & \sbest{77.1} & \sbest{26.4} & \sbest{9.2}  & \sbest{2.3}  & \gbest{38 {\scriptsize$\pm$17}}   & \sbest{6.6k {\scriptsize$\pm$2.1k}} & \gbest{620 {\scriptsize$\pm$197}}  \\
GPT-5.4      & 16.1 {\scriptsize$\pm$19.2}          & 28.2         & 5.1          & 2.7          & 0.6          & \sbest{50 {\scriptsize$\pm$14}}   & 7.6k {\scriptsize$\pm$2.2k}         & \sbest{652 {\scriptsize$\pm$194}}  \\
Sonnet 4.6   & 14.9 {\scriptsize$\pm$16.7}          & 20.4         & 3.6          & 3.6          & 0.0          & 97 {\scriptsize$\pm$25}           & 10.3k {\scriptsize$\pm$3.0k}        & 1.4k {\scriptsize$\pm$573}         \\
Gemini 3 Fl. & \gbest{43.4 {\scriptsize$\pm$25.2}}  & \gbest{77.3} & \gbest{33.9} & \gbest{18.2} & \gbest{4.7}  & 52 {\scriptsize$\pm$12}           & 8.4k {\scriptsize$\pm$2.3k}         & 828 {\scriptsize$\pm$238}           \\
DeepSeek V3.2& 12.7 {\scriptsize$\pm$17.7}          & 19.2         & 5.6          & 2.4          & 0.0          & 219 {\scriptsize$\pm$173}         & 8.2k {\scriptsize$\pm$2.5k}         & 972 {\scriptsize$\pm$371}           \\
Qwen3.5-9B   & 3.9 {\scriptsize$\pm$11.4}           & 5.3          & 1.6          & 1.0          & 0.0          & 534 {\scriptsize$\pm$218}         & \gbest{6.3k {\scriptsize$\pm$1.8k}} & 8.9k {\scriptsize$\pm$3.3k}        \\
\bottomrule
\end{tabular}
\end{table*}

\paragraph{Cross-Backbone Comparison.}
\textit{How do backbones compare in aggregate performance and efficiency?}

Table~\ref{tab:backbone_overall} aggregates all 825 runs per backbone. Gemini achieves the highest pooled satisfaction (43.44\%) and strongest upper-threshold coverage (4.73\% TSR), while GPT-4o is second in satisfaction (39.41\%) but first in efficiency (38.29s runtime, 620 output tokens per run). Qwen is both weakest and most expensive (533.73s, 8895 output tokens). GPT-4o and Gemini are similar at the 20\% threshold (both near 77\%), but Gemini separates at harder thresholds (33.94\% above 50\%, 4.73\% TSR). Extended visualizations appear in Appendix~\ref{app:backbone_comparison_figures}.

\paragraph{Inter-Backbone Agreement.}
We compute per-language Fleiss' $\kappa$ on binary requirement judgments, treating each of 1095 requirement instances as a rated item and the six backbones as raters. Agreement is modest: EN ($\kappa=0.222$), AR (0.184), TR (0.231), ZH (0.184), HI (0.169), with a mean of 0.198. No language exceeds 0.25, confirming that backbone choice affects requirement-level verdicts substantively. Lower agreement in Hindi and Arabic is consistent with stronger multilingual variability in those settings.

\paragraph{Arabic Case Study.}
Arabic provides the clearest example of backbone-dependent language effects: it is the weakest setting under GPT-4o (due to stricter evidence weighting) but the strongest under Gemini (51.72\%, $p<0.001$ vs.\ GPT-4o). The requirement-type breakdown shows that Gemini's advantage spans multiple categories, while GPT-4o drops sharply in semantically dense categories. A detailed analysis is provided in Appendix~\ref{app:arabic_analysis}.

\paragraph{Ablation Study: Prompt Component Localization.}
\textit{Does localizing judge instructions matter beyond translating benchmark content?}

Multilingual Agent-as-a-Judge can localize either only benchmark-facing text or the full judge prompt stack. We study this distinction with a controlled ablation on GPT-4o, fixing the same 20 DevAI tasks, agent trajectories, and backbone, and evaluating Arabic and Hindi across three frameworks (60 runs per language and condition).

In \textit{Full Localization}, both judge-side prompts and benchmark text are in the target language, while in \textit{Partial Localization}, benchmark content remains localized, but judge-side instructions are in English. Table~\ref{tab:ablation_prompt_component_localization} reports pooled satisfaction ($\mu \pm \sigma$) and percentile success rates.

\begin{table*}[t]
\centering
\footnotesize
\renewcommand{\arraystretch}{1.3}
\setlength{\tabcolsep}{5pt}
\caption{Ablation: effect of prompt-component localization (GPT-4o, 20 tasks, 60 runs per row). \textit{Full} = localized system prompts + benchmark text. \textit{Partial} = English judge instructions with localized benchmark text. \colorbox{gold!30}{\strut Gold} = better condition per metric.}
\label{tab:ablation_prompt_component_localization}
\begin{tabular}{@{}ll r rrrr@{}}
\toprule
 & & & \multicolumn{4}{c}{\textbf{Success Thresholds\,(\%)}} \\
\cmidrule(lr){4-7}
\textbf{Lang.} & \textbf{Condition} & \textbf{Sat.\,(\%)} & {$>$20} & {$>$50} & {$>$70} & {100} \\
\midrule
AR & Full    & 23.7 {\scriptsize$\pm$21.5}          & 35.0         & 8.3           & 5.0           & \gbest{3.3} \\
AR & Partial & \gbest{24.1 {\scriptsize$\pm$21.4}}   & 35.0         & \gbest{10.0}  & \gbest{6.7}   & 1.7         \\
\addlinespace[4pt]
HI & Full    & \gbest{42.8 {\scriptsize$\pm$22.3}}   & \gbest{78.3} & \gbest{26.7}  & \gbest{11.7}  & \gbest{5.0} \\
HI & Partial & 23.2 {\scriptsize$\pm$19.3}           & 35.0         & 8.3           & 3.3           & 0.0         \\
\bottomrule
\end{tabular}
\end{table*}

Full localization has uneven effects across languages. For Arabic, results are similar under both conditions (23.65\% vs.\ 24.14\% mean satisfaction; identical 35.00\% above 20\%), with only small differences at higher thresholds and TSR (3.33\% vs.\ 1.67\%). 
In contrast, Hindi shows a large degradation under partial localization: mean satisfaction drops from 42.75\% to 23.19\%, the $>$20\% rate from 78.33\% to 35.00\%, and TSR from 5.00\% to 0.00\%. 

These results indicate that translating benchmark content alone is insufficient in some settings. For Hindi, localized judge instructions improve evaluation stability, whereas Arabic appears less sensitive to instruction language. One possible explanation is that Arabic's stronger representation in LLM training data allows GPT-4o to maintain effective evidence integration even with English-language judge instructions, while Hindi's lower resource level makes the judge more dependent on target-language instructions for consistent reasoning. Prompt localization should therefore be treated as an explicit experimental factor rather than an implementation detail.

\section{Discussion}

Three findings emerge from the experiments. First, backbone ranking is language-dependent: GPT-4o leads in English, while Gemini leads in Arabic and Hindi. This means that any evaluation conducted in a single language risks producing misleading backbone rankings. Second, backbone capability gates whether multilingual effects are observable at all: Qwen and DeepSeek show negligible cross-language variation because their baseline performance is too low for language to make a measurable difference. Third, the type of requirement matters: semantically dense categories (model construction, evaluation metrics) are more sensitive to language than operationally concrete categories (data loading, training), at least among stronger backbones.

\paragraph{Token-Length Confound.}
A natural alternative explanation for the cross-language differences is that languages with longer tokenized prompts produce systematically different judge behavior. We tested this by computing per-backbone Spearman correlations between language-level mean input tokens and language-level mean satisfaction. The correlations are inconsistent in sign and non-significant for all six backbones: GPT-4o ($\rho=-0.70$, $p=0.188$), GPT-5.4 ($\rho=0.60$, $p=0.285$), Sonnet ($\rho=0.10$, $p=0.873$), Gemini ($\rho=0.30$, $p=0.624$), DeepSeek ($\rho=0.30$, $p=0.624$), and Qwen ($\rho=0.50$, $p=0.391$). The current evidence does not support a tokenization-only account of the multilingual effects. However, with only five languages per correlation, statistical power is limited, so a token-length contribution cannot be definitively ruled out.

\paragraph{Validity of Judge Outputs.}
Without a multilingual human baseline, a natural concern is whether the observed backbone-language interactions reflect genuine evaluation differences or systematic judging artifacts. Several features of the data argue against the artifact interpretation. The backbone-ranking inversion (GPT-4o strongest in English, Gemini strongest in Arabic) would only be an artifact if all six backbones were systematically wrong in the same backbone-specific, language-specific way across thousands of individual requirement judgments. The modest inter-backbone agreement (Fleiss' $\kappa < 0.25$) shows that backbones disagree substantially on which requirements are satisfied, which is inconsistent with a shared systematic bias. Furthermore, the requirement-type analysis reveals that language sensitivity is concentrated in semantically dense categories rather than uniformly distributed, which is more consistent with genuine interpretation difficulty than with random or uniform artifacts. While these observations do not substitute for human validation, they increase confidence that the observed interactions reflect real evaluation behavior rather than noise.

\paragraph{Practical Recommendations.}
Three reporting norms follow from our results. First, backbone identity and evaluation language should be reported as first-class metadata, because the same benchmark can produce different rankings under different judge configurations. Second, non-English evaluation should be validated with at least two judge backbones, since several cross-language conclusions in this study change when the backbone changes. Third, percentile success profiles should accompany TSR, because the 100\% completion threshold hides the broader region of meaningful partial progress revealed by lower thresholds. Framework-resolved breakdowns, which may reveal additional backbone-language-framework interactions, are provided in the supplementary material.

\section{Conclusion}

This study demonstrates that multilingual prompt localization changes judge behavior in backbone-dependent ways that cannot be predicted from English-only evaluation. The central finding is that backbone rankings invert across languages: GPT-4o leads in English, while Gemini leads in Arabic and Hindi. This interaction, combined with modest inter-backbone agreement (Fleiss' $\kappa < 0.25$ in all languages) and the ablation showing that instruction-language localization can be decisive (Hindi satisfaction drops from 42.75\% to 23.19\% under partial localization), indicates that language is not a neutral evaluation parameter.
Across all settings, requirement-level evaluation remains more informative than binary task completion, with low TSR despite substantial partial progress \citep{zhuge2024agentjudge}. This reinforces the value of process-aware benchmarks for realistic development tasks. Future work should extend these findings with additional languages, human alignment baselines, and cross-backbone calibration studies.

\section{Limitations}
Several constraints in our setup limit the scope of our conclusions.


\textbf{Multilingual human alignment.}
We evaluate multilingual prompt localization through judge outputs rather than new multilingual human annotation, so alignment with expert consensus—especially for nuanced or partially satisfied requirements—cannot be fully verified. However, our design compares the \textit{relative} effect of language within each backbone (same tasks, pipeline, and artifacts), isolating sensitivity even without absolute alignment. A comprehensive human validation study is in progress.

\textbf{Self-Termination reporting.}
Self-Termination cannot be analyzed reliably because the necessary termination metadata is not consistently exposed in the saved multilingual outputs. This limits our ability to study stopping behavior and failure modes across languages.

\textbf{Cost analysis.}
The benchmark package provides zero-valued cost fields, which prevents a multilingual cost analysis comparable to that in the original benchmark. We report runtime and token statistics as partial proxies but cannot assess monetary costs.

\textbf{Model and setup scope.}
Our findings are limited to the six judge backbones, prompt stack, and DevAI benchmark configuration used in this study. Different models, prompting strategies, or evaluation pipelines may exhibit different multilingual behavior.

\textbf{Language and framework coverage.}
We evaluate five languages and three open-source frameworks chosen for typological diversity, but this coverage remains limited. Broader claims about multilingual robustness should be interpreted as directional rather than exhaustive.


\bibliographystyle{colm2026_conference}

\begin{thebibliography}{28}
\providecommand{\natexlab}[1]{#1}
\providecommand{\url}[1]{\texttt{#1}}
\expandafter\ifx\csname urlstyle\endcsname\relax
  \providecommand{\doi}[1]{doi: #1}\else
  \providecommand{\doi}{doi: \begingroup \urlstyle{rm}\Url}\fi

\bibitem[Austin et~al.(2021)Austin, Odena, Nye, Bosma, Michalewski, Dohan, Jiang, Cai, Terry, Le, et~al.]{austin2021program}
Jacob Austin, Augustus Odena, Maxwell Nye, Maarten Bosma, Henryk Michalewski, David Dohan, Ellen Jiang, Carrie Cai, Michael Terry, Quoc Le, et~al.
\newblock Program synthesis with large language models.
\newblock \emph{arXiv preprint arXiv:2108.07732}, 2021.

\bibitem[Cassano et~al.(2023)Cassano, Gouwar, Nguyen, Nguyen, Phipps-Costin, Pinckney, Yee, Zi, Anderson, Feldman, et~al.]{cassano2023multipl}
Federico Cassano, Jacob Gouwar, Daniel Nguyen, Sydney Nguyen, Luna Phipps-Costin, Donald Pinckney, Ming-Ho Yee, Yangtian Zi, Carolyn~Jane Anderson, Molly~Q. Feldman, et~al.
\newblock {MultiPL-E}: A scalable and polyglot approach to benchmarking neural code generation.
\newblock \emph{IEEE Transactions on Software Engineering}, 49\penalty0 (7):\penalty0 3675--3691, 2023.

\bibitem[Chai et~al.(2025)Chai, Liu, Yang, Yin, Jin, Liu, Sun, Zhang, Ren, Guo, et~al.]{chai2025mceval}
Linzheng Chai, Shukai Liu, Jian Yang, Yuwei Yin, Ke~Jin, Jiaheng Liu, Tao Sun, Ge~Zhang, Changyu Ren, Hongcheng Guo, et~al.
\newblock {McEval}: Massively multilingual code evaluation.
\newblock In \emph{International Conference on Learning Representations}, 2025.

\bibitem[Chan et~al.(2024)Chan, Chowdhury, Jaffe, Aung, Sherburn, Mays, Starace, Liu, Maksin, Patwardhan, Weng, et~al.]{chan2024mlebench}
Jun~Shern Chan, Neil Chowdhury, Oliver Jaffe, James Aung, Dane Sherburn, Evan Mays, Giulio Starace, Kevin Liu, Leon Maksin, Tejal~A. Patwardhan, Lilian Weng, et~al.
\newblock {MLE-Bench}: Evaluating machine learning agents on machine learning engineering.
\newblock \emph{arXiv preprint arXiv:2410.07095}, 2024.

\bibitem[Chen et~al.(2021)Chen, Tworek, Jun, Yuan, Pinto, Kaplan, Edwards, Burda, Joseph, Brockman, et~al.]{chen2021codex}
Mark Chen, Jerry Tworek, Heewoo Jun, Qiming Yuan, Henrique Ponde de~Oliveira Pinto, Jared Kaplan, Harri Edwards, Yuri Burda, Nicholas Joseph, Greg Brockman, et~al.
\newblock Evaluating large language models trained on code.
\newblock \emph{arXiv preprint arXiv:2107.03374}, 2021.

\bibitem[Doddapaneni et~al.(2025)Doddapaneni, Khan, Venkatesh, Dabre, Kunchukuttan, and Khapra]{doddapaneni2025cross}
Sumanth Doddapaneni, Mohammed Safi Ur~Rahman Khan, Dilip Venkatesh, Raj Dabre, Anoop Kunchukuttan, and Mitesh~M. Khapra.
\newblock Cross-lingual auto evaluation for assessing multilingual {LLM}s.
\newblock In \emph{Proceedings of the 63rd Annual Meeting of the Association for Computational Linguistics (Volume 1: Long Papers)}, pp.\  29297--29329, Vienna, Austria, 2025. Association for Computational Linguistics.
\newblock \doi{10.18653/v1/2025.acl-long.1419}.
\newblock URL \url{https://aclanthology.org/2025.acl-long.1419/}.

\bibitem[FitzGerald et~al.(2023)FitzGerald, Hench, Peris, Mackie, Rottmann, Sanchez, Nash, Urbach, Kakarala, Singh, Ranganath, Crist, Britan, Leeuwis, Tur, and Natarajan]{fitzgerald2023massive}
Jack FitzGerald, Christopher Hench, Charith Peris, Scott Mackie, Kay Rottmann, Ana Sanchez, Aaron Nash, Liam Urbach, Vishesh Kakarala, Richa Singh, Swetha Ranganath, Laurie Crist, Misha Britan, Wouter Leeuwis, Gokhan Tur, and Prem Natarajan.
\newblock {MASSIVE}: A 1m-example multilingual natural language understanding dataset with 51 typologically-diverse languages.
\newblock In \emph{Proceedings of the 61st Annual Meeting of the Association for Computational Linguistics (Volume 1: Long Papers)}, pp.\  4277--4302, Toronto, Canada, 2023. Association for Computational Linguistics.
\newblock \doi{10.18653/v1/2023.acl-long.235}.
\newblock URL \url{https://aclanthology.org/2023.acl-long.235/}.

\bibitem[Fu \& Liu(2025)Fu and Liu]{fu2025reliable}
Xiyan Fu and Wei Liu.
\newblock How reliable is multilingual {LLM}-as-a-judge?
\newblock In \emph{Findings of the Association for Computational Linguistics: {EMNLP} 2025}, pp.\  11040--11053, Suzhou, China, 2025. Association for Computational Linguistics.
\newblock \doi{10.18653/v1/2025.findings-emnlp.587}.
\newblock URL \url{https://aclanthology.org/2025.findings-emnlp.587/}.

\bibitem[Gururangan et~al.(2018)Gururangan, Swayamdipta, Levy, Schwartz, Bowman, and Smith]{gururangan2018annotation}
Suchin Gururangan, Swabha Swayamdipta, Omer Levy, Roy Schwartz, Samuel Bowman, and Noah~A Smith.
\newblock Annotation artifacts in natural language inference data.
\newblock In \emph{Proceedings of the 2018 Conference of the North American Chapter of the Association for Computational Linguistics: Human Language Technologies, Volume 2 (Short Papers)}, pp.\  107--112, 2018.

\bibitem[Habash(2010)]{habash2010arabic}
Nizar~Y. Habash.
\newblock \emph{Introduction to Arabic Natural Language Processing}, volume~10 of \emph{Synthesis Lectures on Human Language Technologies}.
\newblock Morgan \& Claypool, 2010.

\bibitem[Hong et~al.(2024)Hong, Zhuge, Chen, Zheng, Cheng, Wang, Zhang, Wang, Yau, Lin, et~al.]{hong2024metagpt}
Sirui Hong, Mingchen Zhuge, Jonathan Chen, Xiawu Zheng, Yuheng Cheng, Jinlin Wang, Ceyao Zhang, Zili Wang, Steven Ka~Shing Yau, Zijuan Lin, et~al.
\newblock {MetaGPT}: Meta programming for a multi-agent collaborative framework.
\newblock In \emph{The Twelfth International Conference on Learning Representations}, 2024.

\bibitem[Huang \& Zhao(2006)Huang and Zhao]{huang2006chinese}
Chang-Ning Huang and Hai Zhao.
\newblock Which is essential for chinese word segmentation: Character versus word.
\newblock In \emph{Proceedings of the 20th Pacific Asia Conference on Language, Information and Computation}, pp.\  1--12, Huazhong Normal University, Wuhan, China, 2006. Tsinghua University Press.
\newblock URL \url{https://aclanthology.org/Y06-1001/}.

\bibitem[Huang et~al.(2025)Huang, Vishwakarma, and Sala]{huang2025pajama}
Tzu-Heng Huang, Harit Vishwakarma, and Frederic Sala.
\newblock Time to impeach {LLM}-as-a-judge: Programs are the future of evaluation.
\newblock \emph{arXiv preprint arXiv:2506.10403}, 2025.

\bibitem[Jimenez et~al.(2023)Jimenez, Yang, Wettig, Lieret, Yao, Pei, Press, and Narasimhan]{jimenez2023swebench}
Carlos~E. Jimenez, John Yang, Alexander Wettig, Kilian Lieret, Shunyu Yao, Kexin Pei, Ofir Press, and Karthik Narasimhan.
\newblock {SWE-Bench}: Can language models resolve real-world {GitHub} issues?
\newblock \emph{arXiv preprint arXiv:2310.06770}, 2023.

\bibitem[Magar \& Schwartz(2022)Magar and Schwartz]{magar2022data}
Inbal Magar and Roy Schwartz.
\newblock Data contamination: From memorization to exploitation.
\newblock In \emph{Proceedings of the 60th Annual Meeting of the Association for Computational Linguistics (Volume 2: Short Papers)}, pp.\  157--165, 2022.

\bibitem[Ni et~al.(2025)Ni, Wang, Zhang, Lu, He, You, Tang, Du, Sun, Liu, et~al.]{ni2025gittaskbench}
Ziyi Ni, Huacan Wang, Shuo Zhang, Shuo Lu, Ziyang He, Wang You, Zhenheng Tang, Yuntao Du, Bill Sun, Hongzhang Liu, et~al.
\newblock {GitTaskBench}: A benchmark for code agents solving real-world tasks through code repository leveraging.
\newblock \emph{arXiv preprint arXiv:2508.18993}, 2025.

\bibitem[Ozturel et~al.(2019)Ozturel, Kayadelen, and Demirsahin]{ozturel2019turkish}
Adnan Ozturel, Tolga Kayadelen, and Isin Demirsahin.
\newblock A syntactically expressive morphological analyzer for turkish.
\newblock In \emph{Proceedings of the 14th International Conference on Finite-State Methods and Natural Language Processing}, pp.\  65--75, Dresden, Germany, 2019. Association for Computational Linguistics.
\newblock \doi{10.18653/v1/W19-3110}.
\newblock URL \url{https://aclanthology.org/W19-3110/}.

\bibitem[{Pythagora.io}(2023)]{pythagora2023gptpilot}
{Pythagora.io}.
\newblock {GPT-Pilot}: Your {AI} copilot for software development.
\newblock GitHub repository, 2023.
\newblock URL \url{https://github.com/Pythagora-io/gpt-pilot}.

\bibitem[Raihan et~al.(2025)Raihan, Anastasopoulos, and Zampieri]{raihan2025mhumaneval}
Nishat Raihan, Antonios Anastasopoulos, and Marcos Zampieri.
\newblock {mHumanEval}: A multilingual benchmark to evaluate large language models for code generation.
\newblock In \emph{Proceedings of the 2025 Conference of the North American Chapter of the Association for Computational Linguistics}, 2025.

\bibitem[Sheth et~al.(2026)Sheth, Jonke, Mantrach, and Mansour]{sheth2026crosslingual}
Ivaxi Sheth, Zeno Jonke, Amin Mantrach, and Saab Mansour.
\newblock Cross-lingual {LLM}-judge transfer via evaluation decomposition.
\newblock \emph{arXiv preprint arXiv:2603.18557}, 2026.
\newblock \doi{10.48550/arXiv.2603.18557}.
\newblock URL \url{https://arxiv.org/abs/2603.18557}.

\bibitem[Singh et~al.(2024)Singh, Gupta, Bharadwaj, Tewari, and Talukdar]{singh2024indicgenbench}
Harman Singh, Nitish Gupta, Shikhar Bharadwaj, Dinesh Tewari, and Partha Talukdar.
\newblock {IndicGenBench}: A multilingual benchmark to evaluate generation capabilities of {LLM}s on indic languages.
\newblock In \emph{Proceedings of the 62nd Annual Meeting of the Association for Computational Linguistics (Volume 1: Long Papers)}, pp.\  11047--11073, Bangkok, Thailand, 2024. Association for Computational Linguistics.
\newblock \doi{10.18653/v1/2024.acl-long.595}.
\newblock URL \url{https://aclanthology.org/2024.acl-long.595/}.

\bibitem[Son et~al.(2024)Son, Yoon, Suk, Aula-Blasco, Aslan, Kim, Islam, Prats-Cristi{\`a}, Tormo-Ba{\~n}uelos, and Kim]{son2024mmeval}
Guijin Son, Dongkeun Yoon, Juyoung Suk, Javier Aula-Blasco, Mano Aslan, Vu~Trong Kim, Shayekh~Bin Islam, Jaume Prats-Cristi{\`a}, Luc{\'i}a Tormo-Ba{\~n}uelos, and Seungone Kim.
\newblock {MM-Eval}: A multilingual meta-evaluation benchmark for {LLM}-as-a-judge and reward models.
\newblock \emph{arXiv preprint arXiv:2410.17578}, 2024.
\newblock \doi{10.48550/arXiv.2410.17578}.
\newblock URL \url{https://arxiv.org/abs/2410.17578}.

\bibitem[Sonwane et~al.(2026)Sonwane, Tu, Lu, Beger, Larsen, Dhar, Alford, Chen, Pattanayak, Dang, et~al.]{sonwane2026omnicode}
Atharv Sonwane, Eng-Shen Tu, Wei-Chung Lu, Claas Beger, Carter Larsen, Debjit Dhar, Simon Alford, Rachel Chen, Ronit Pattanayak, Tuan~Anh Dang, et~al.
\newblock {OmniCode}: A benchmark for evaluating software engineering agents.
\newblock \emph{arXiv preprint arXiv:2602.02262}, 2026.

\bibitem[Wang et~al.(2024)Wang, Li, Song, Xu, Tang, Zhuge, Pan, Song, Li, Singh, et~al.]{wang2024opendevin}
Xingyao Wang, Boxuan Li, Yufan Song, Frank~F. Xu, Xiangru Tang, Mingchen Zhuge, Jiayi Pan, Yueqi Song, Bowen Li, Jaskirat Singh, et~al.
\newblock {OpenDevin}: An open platform for {AI} software developers as generalist agents.
\newblock \emph{arXiv preprint arXiv:2407.16741}, 2024.

\bibitem[Xu et~al.(2025)Xu, Deng, Li, Yu, Tang, Huang, Lai, Zhan, Wu, Zhang, et~al.]{xu2025swecompass}
Jingxuan Xu, Ken Deng, Weihao Li, Songwei Yu, Huaixi Tang, Haoyang Huang, Zhiyi Lai, Zizheng Zhan, Yanan Wu, Chenchen Zhang, et~al.
\newblock {SWE-Compass}: Towards unified evaluation of agentic coding abilities for large language models.
\newblock \emph{arXiv preprint arXiv:2511.05459}, 2025.
\newblock \doi{10.48550/arXiv.2511.05459}.
\newblock URL \url{https://arxiv.org/abs/2511.05459}.

\bibitem[Ye et~al.(2024)Ye, Kim, Kim, Hwang, Kim, Jo, Thorne, Kim, and Seo]{ye2024flask}
Seonghyeon Ye, Doyoung Kim, Sungdong Kim, Hyeonbin Hwang, Seungone Kim, Yongrae Jo, James Thorne, Juho Kim, and Minjoon Seo.
\newblock {FLASK}: Fine-grained language model evaluation based on alignment skill sets.
\newblock In \emph{The Twelfth International Conference on Learning Representations}, 2024.

\bibitem[Zheng et~al.(2024)Zheng, Chiang, Sheng, Zhuang, Wu, Zhuang, Lin, Li, Li, Xing, et~al.]{zheng2024judging}
Lianmin Zheng, Wei-Lin Chiang, Siyuan Sheng, Sihan Zhuang, Zhanghao Wu, Yonghao Zhuang, Zi~Lin, Zhuohan Li, Dacheng Li, Eric~P. Xing, et~al.
\newblock Judging {LLM}-as-a-judge with {MT-Bench} and {Chatbot Arena}.
\newblock In \emph{Advances in Neural Information Processing Systems}, 2024.

\bibitem[Zhuge et~al.(2024)Zhuge, Zhao, Ashley, Wang, Khizbullin, Xiong, Liu, Chang, Krishnamoorthi, Tian, et~al.]{zhuge2024agentjudge}
Mingchen Zhuge, Changsheng Zhao, Dylan~R. Ashley, Wenyi Wang, Dmitrii Khizbullin, Yunyang Xiong, Zechun Liu, Ernie Chang, Raghuraman Krishnamoorthi, Yuandong Tian, et~al.
\newblock Agent-as-a-judge: Evaluate agents with agents.
\newblock \emph{arXiv preprint arXiv:2410.10934}, 2024.

\end{thebibliography}

\clearpage
\appendix
\onecolumn
\section{Extended Backbone Comparison}
\label{app:backbone_comparison_figures}

\begin{figure}[t]
\centering
\includegraphics[width=\linewidth]{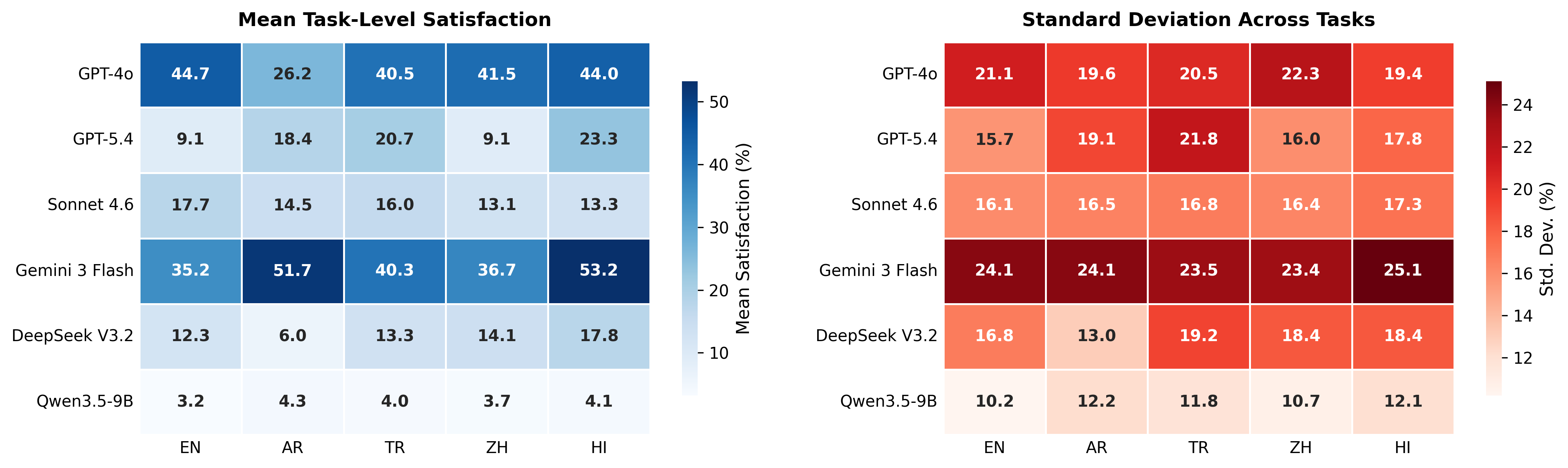}
\caption{Cross-backbone heatmaps summarizing central tendency and variability, averaged over three developer frameworks. \textbf{Left:} Mean task-level requirement satisfaction (\%), using a sequential blue colormap (darker = higher). \textbf{Right:} Standard deviation across tasks, using a sequential red colormap (darker = more variable). Reading both panels together reveals whether a strong backbone-language combination is also stable or whether its gains come with large task-level dispersion.}
\label{fig:backbone_satisfaction_variance_heatmaps}
\end{figure}

\begin{figure}[t]
\centering
\includegraphics[width=\linewidth]{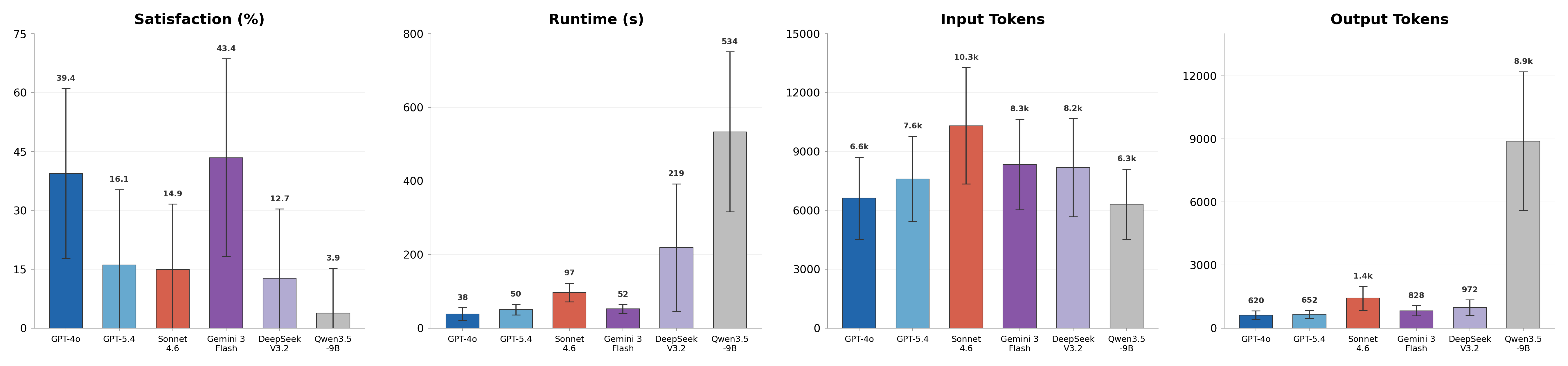}
\caption{Cross-backbone performance and cost with standard-deviation error bars (825 runs per backbone). GPT-4o achieves the best efficiency profile; Qwen is both weakest and most expensive.}
\label{fig:backbone_overall_error_bars}
\end{figure}

Figures~\ref{fig:backbone_satisfaction_variance_heatmaps} and~\ref{fig:backbone_overall_error_bars} complement the main-text backbone analysis by decomposing aggregate statistics into their variance components.

\section{Significance Tests}
\label{app:significance_tests} We report exploratory significance analyses for the main satisfaction comparisons. Each test uses paired task-level satisfaction scores over the same 55 DevAI tasks after averaging over the three developer-agent frameworks, applies a two-sided Wilcoxon signed-rank test, and reports a 95\% bootstrap confidence interval for the paired median difference.

\subsection{Cross-language comparisons within each backbone}
\noindent Two-sided Wilcoxon signed-rank tests over paired task-level satisfaction scores (55 paired DevAI tasks after averaging over the three developer-agent frameworks). $\Delta$ denotes right minus left, and the confidence interval is a 95\% bootstrap interval for the paired median difference. We report both raw $p$-values and Holm-Bonferroni-adjusted $p_{\mathrm{Holm}}$ values after correcting over all pairwise Wilcoxon tests reported in this appendix.

\begingroup
\renewcommand{\arraystretch}{1.08}
\setlength{\tabcolsep}{3.5pt}
\footnotesize
\begin{longtable}{@{}l ll r ll r p{0.12\linewidth} rl@{}}
\caption{Pairwise language comparisons per judge backbone.
$\Delta$ = median difference; CI = bootstrap 95\% confidence interval.
Significance after Holm correction:
{$^{***}$}\,$p_{\mathrm{Holm}}\!<\!0.001$;~
{$^{**}$}\,$p_{\mathrm{Holm}}\!<\!0.01$;~
{$^{*}$}\,$p_{\mathrm{Holm}}\!<\!0.05$;~
n.s.\ otherwise.}
\label{tab:pairwise_language}\\
\toprule
\textbf{Backbone} & \textbf{Left} & \textbf{Mean\textsubscript{L}} & \textbf{Right} & \textbf{Mean\textsubscript{R}} & \textbf{$\Delta$} & \textbf{95\% CI} & \textbf{$p$} & \textbf{$p_{\mathrm{Holm}}$} & \\
\midrule
\endfirsthead
\toprule
\textbf{Backbone} & \textbf{Left} & \textbf{Mean\textsubscript{L}} & \textbf{Right} & \textbf{Mean\textsubscript{R}} & \textbf{$\Delta$} & \textbf{95\% CI} & \textbf{$p$} & \textbf{$p_{\mathrm{Holm}}$} & \\
\midrule
\endhead
\midrule
\multicolumn{10}{r}{\textit{Continued on next page}} \\
\endfoot
\bottomrule
\endlastfoot
%
GPT-4o & EN & 44.72 & AR & 26.25 & $-$19.05 & [$-$23.81, $-$14.29] & $<$0.001 & $<$\textbf{0.001} & $^{***}$ \\
GPT-4o & EN & 44.72 & TR & 40.53 & $-$4.76  & [$-$12.50, 4.76]     & 0.053    & 0.952             &          \\
GPT-4o & EN & 44.72 & ZH & 41.53 & +0.00    & [$-$8.33, 4.17]      & 0.193    & 0.952             &          \\
GPT-4o & EN & 44.72 & HI & 44.02 & +0.00    & [$-$5.56, 4.76]      & 0.637    & 0.952             &          \\
GPT-4o & AR & 26.25 & TR & 40.53 & +13.33   & [9.52, 19.05]        & $<$0.001 & $<$\textbf{0.001} & $^{***}$ \\
GPT-4o & AR & 26.25 & ZH & 41.53 & +14.29   & [9.52, 19.05]        & $<$0.001 & $<$\textbf{0.001} & $^{***}$ \\
GPT-4o & AR & 26.25 & HI & 44.02 & +19.05   & [14.29, 22.22]       & $<$0.001 & $<$\textbf{0.001} & $^{***}$ \\
GPT-4o & TR & 40.53 & ZH & 41.53 & +2.38    & [0.00, 5.56]         & 0.739    & 0.952             &          \\
GPT-4o & TR & 40.53 & HI & 44.02 & +4.76    & [0.00, 9.52]         & 0.122    & 0.952             &          \\
GPT-4o & ZH & 41.53 & HI & 44.02 & +4.76    & [0.00, 9.52]         & 0.111    & 0.952             &          \\
\midrule
%
GPT-5.4 & EN & 9.09  & AR & 18.40 & +6.67    & [4.76, 9.52]         & $<$0.001 & $<$\textbf{0.001} & $^{***}$ \\
GPT-5.4 & EN & 9.09  & TR & 20.67 & +9.52    & [4.76, 13.33]        & $<$0.001 & $<$\textbf{0.001} & $^{***}$ \\
GPT-5.4 & EN & 9.09  & ZH &  9.05 & +0.00    & [0.00, 0.00]         & 0.952    & 0.952             &          \\
GPT-5.4 & EN & 9.09  & HI & 23.30 & +13.33   & [9.52, 16.67]        & $<$0.001 & $<$\textbf{0.001} & $^{***}$ \\
GPT-5.4 & AR & 18.40 & TR & 20.67 & +0.00    & [0.00, 4.76]         & 0.080    & 0.952             &          \\
GPT-5.4 & AR & 18.40 & ZH &  9.05 & $-$8.33  & [$-$9.52, $-$4.76]   & $<$0.001 & $<$\textbf{0.001} & $^{***}$ \\
GPT-5.4 & AR & 18.40 & HI & 23.30 & +4.76    & [3.70, 6.67]         & 0.001    & \textbf{0.048}    & $^{*}$   \\
GPT-5.4 & TR & 20.67 & ZH &  9.05 & $-$9.52  & [$-$14.29, $-$5.56]  & $<$0.001 & $<$\textbf{0.001} & $^{***}$ \\
GPT-5.4 & TR & 20.67 & HI & 23.30 & +0.00    & [0.00, 4.76]         & 0.024    & 0.849             &          \\
GPT-5.4 & ZH &  9.05 & HI & 23.30 & +14.29   & [9.52, 19.05]        & $<$0.001 & $<$\textbf{0.001} & $^{***}$ \\
\midrule
%
Son.\ 4.6 & EN & 17.68 & AR & 14.53 & +0.00  & [0.00, 0.00]         & $<$0.001 & \textbf{0.032}    & $^{*}$   \\
Son.\ 4.6 & EN & 17.68 & TR & 15.99 & +0.00  & [0.00, 0.00]         & 0.015    & 0.583             &          \\
Son.\ 4.6 & EN & 17.68 & ZH & 13.14 & +0.00  & [$-$4.76, 0.00]      & $<$0.001 & \textbf{0.003}    & $^{**}$  \\
Son.\ 4.6 & EN & 17.68 & HI & 13.30 & +0.00  & [$-$4.76, 0.00]      & $<$0.001 & \textbf{0.005}    & $^{**}$  \\
Son.\ 4.6 & AR & 14.53 & TR & 15.99 & +0.00  & [0.00, 0.00]         & 0.030    & 0.952             &          \\
Son.\ 4.6 & AR & 14.53 & ZH & 13.14 & +0.00  & [0.00, 0.00]         & 0.218    & 0.952             &          \\
Son.\ 4.6 & AR & 14.53 & HI & 13.30 & +0.00  & [0.00, 0.00]         & 0.131    & 0.952             &          \\
Son.\ 4.6 & TR & 15.99 & ZH & 13.14 & +0.00  & [0.00, 0.00]         & 0.005    & 0.231             &          \\
Son.\ 4.6 & TR & 15.99 & HI & 13.30 & +0.00  & [0.00, 0.00]         & 0.003    & 0.116             &          \\
Son.\ 4.6 & ZH & 13.14 & HI & 13.30 & +0.00  & [0.00, 0.00]         & 0.931    & 0.952             &          \\
\midrule
%
Gem.\ 3 Fl. & EN & 35.21 & AR & 51.72 & +16.67   & [11.11, 20.00]       & $<$0.001 & $<$\textbf{0.001} & $^{***}$ \\
Gem.\ 3 Fl. & EN & 35.21 & TR & 40.30 & +4.76    & [0.00, 9.52]         & $<$0.001 & \textbf{0.011}    & $^{*}$   \\
Gem.\ 3 Fl. & EN & 35.21 & ZH & 36.72 & +0.00    & [0.00, 4.76]         & 0.143    & 0.952             &          \\
Gem.\ 3 Fl. & EN & 35.21 & HI & 53.22 & +19.05   & [14.29, 22.22]       & $<$0.001 & $<$\textbf{0.001} & $^{***}$ \\
Gem.\ 3 Fl. & AR & 51.72 & TR & 40.30 & $-$9.52  & [$-$14.29, $-$5.56]  & $<$0.001 & $<$\textbf{0.001} & $^{***}$ \\
Gem.\ 3 Fl. & AR & 51.72 & ZH & 36.72 & $-$9.52  & [$-$14.29, $-$9.52]  & $<$0.001 & $<$\textbf{0.001} & $^{***}$ \\
Gem.\ 3 Fl. & AR & 51.72 & HI & 53.22 & +0.00    & [$-$4.76, 5.56]      & 0.385    & 0.952             &          \\
Gem.\ 3 Fl. & TR & 40.30 & ZH & 36.72 & $-$4.76  & [$-$6.67, 0.00]      & 0.026    & 0.903             &          \\
Gem.\ 3 Fl. & TR & 40.30 & HI & 53.22 & +14.29   & [6.67, 19.05]        & $<$0.001 & $<$\textbf{0.001} & $^{***}$ \\
Gem.\ 3 Fl. & ZH & 36.72 & HI & 53.22 & +14.29   & [13.33, 20.00]       & $<$0.001 & $<$\textbf{0.001} & $^{***}$ \\
\midrule
%
DeepSeek & EN & 12.29 & AR &  6.02 & $-$4.76  & [$-$9.52, 0.00]      & $<$0.001 & $<$\textbf{0.001} & $^{***}$ \\
DeepSeek & EN & 12.29 & TR & 13.33 & +0.00    & [0.00, 4.76]         & 0.293    & 0.952             &          \\
DeepSeek & EN & 12.29 & ZH & 14.07 & +0.00    & [0.00, 4.17]         & 0.139    & 0.952             &          \\
DeepSeek & EN & 12.29 & HI & 17.77 & +5.56    & [0.00, 9.52]         & $<$0.001 & \textbf{0.008}    & $^{**}$  \\
DeepSeek & AR &  6.02 & TR & 13.33 & +0.00    & [0.00, 5.56]         & $<$0.001 & $<$\textbf{0.001} & $^{***}$ \\
DeepSeek & AR &  6.02 & ZH & 14.07 & +4.76    & [4.76, 9.52]         & $<$0.001 & $<$\textbf{0.001} & $^{***}$ \\
DeepSeek & AR &  6.02 & HI & 17.77 & +11.11   & [5.56, 14.29]        & $<$0.001 & $<$\textbf{0.001} & $^{***}$ \\
DeepSeek & TR & 13.33 & ZH & 14.07 & +0.00    & [0.00, 0.00]         & 0.816    & 0.952             &          \\
DeepSeek & TR & 13.33 & HI & 17.77 & +4.76    & [4.76, 5.56]         & $<$0.001 & \textbf{0.014}    & $^{*}$   \\
DeepSeek & ZH & 14.07 & HI & 17.77 & +4.76    & [0.00, 5.56]         & 0.004    & 0.169             &          \\
\midrule
%
Qwen 9B & EN &  3.20 & AR &  4.32 & +0.00    & [0.00, 0.00]         & 0.031    & 0.952             &          \\
Qwen 9B & EN &  3.20 & TR &  3.95 & +0.00    & [0.00, 0.00]         & 0.234    & 0.952             &          \\
Qwen 9B & EN &  3.20 & ZH &  3.66 & +0.00    & [0.00, 0.00]         & 0.125    & 0.952             &          \\
Qwen 9B & EN &  3.20 & HI &  4.11 & +0.00    & [0.00, 0.00]         & 0.172    & 0.952             &          \\
Qwen 9B & AR &  4.32 & TR &  3.95 & +0.00    & [0.00, 0.00]         & 0.344    & 0.952             &          \\
Qwen 9B & AR &  4.32 & ZH &  3.66 & +0.00    & [0.00, 0.00]         & 0.109    & 0.952             &          \\
Qwen 9B & AR &  4.32 & HI &  4.11 & +0.00    & [0.00, 0.00]         & 0.676    & 0.952             &          \\
Qwen 9B & TR &  3.95 & ZH &  3.66 & +0.00    & [0.00, 0.00]         & 0.562    & 0.952             &          \\
Qwen 9B & TR &  3.95 & HI &  4.11 & +0.00    & [0.00, 0.00]         & 0.625    & 0.952             &          \\
Qwen 9B & ZH &  3.66 & HI &  4.11 & +0.00    & [0.00, 0.00]         & 0.312    & 0.952             &          \\
\end{longtable}
\endgroup

\subsection{Cross-backbone comparisons within each language}
\noindent Two-sided Wilcoxon signed-rank tests over paired task-level satisfaction scores (55 paired DevAI tasks after averaging over the three developer-agent frameworks). $\Delta$ denotes right minus left, and the confidence interval is a 95\% bootstrap interval for the paired median difference. We report both raw $p$-values and Holm-Bonferroni-adjusted $p_{\mathrm{Holm}}$ values after correcting over all pairwise Wilcoxon tests reported in this appendix.

\begingroup
\renewcommand{\arraystretch}{1.08}
\setlength{\tabcolsep}{3.5pt}
\footnotesize
\begin{longtable}{@{}l ll r ll r p{0.12\linewidth} rl@{}}
\caption{Pairwise backbone comparisons per language.
$\Delta$ = median difference; CI = bootstrap 95\% confidence interval.
Significance after Holm correction:
{$^{***}$}\,$p_{\mathrm{Holm}}\!<\!0.001$;~
{$^{**}$}\,$p_{\mathrm{Holm}}\!<\!0.01$;~
{$^{*}$}\,$p_{\mathrm{Holm}}\!<\!0.05$;~
n.s.\ otherwise.}
\label{tab:pairwise_backbone}\\
\toprule
\textbf{Lang.} & \textbf{Left} & \textbf{Mean\textsubscript{L}} & \textbf{Right} & \textbf{Mean\textsubscript{R}} & \textbf{$\Delta$} & \textbf{95\% CI} & \textbf{$p$} & \textbf{$p_{\mathrm{Holm}}$} & \\
\midrule
\endfirsthead
\toprule
\textbf{Lang.} & \textbf{Left} & \textbf{Mean\textsubscript{L}} & \textbf{Right} & \textbf{Mean\textsubscript{R}} & \textbf{$\Delta$} & \textbf{95\% CI} & \textbf{$p$} & \textbf{$p_{\mathrm{Holm}}$} & \\
\midrule
\endhead
\midrule
\multicolumn{10}{r}{\textit{Continued on next page}} \\
\endfoot
\bottomrule
\endlastfoot
%
EN & GPT-4o    & 44.72 & GPT-5.4     &  9.09 & $-$33.33 & [$-$38.10, $-$33.33] & $<$0.001 & $<$\textbf{0.001} & $^{***}$ \\
EN & GPT-4o    & 44.72 & Son.\ 4.6   & 17.68 & $-$25.93 & [$-$33.33, $-$23.81] & $<$0.001 & $<$\textbf{0.001} & $^{***}$ \\
EN & GPT-4o    & 44.72 & Gem.\ 3 Fl. & 35.21 & $-$9.52  & [$-$19.05, $-$4.76]  & $<$0.001 & 0.036             & $^{*}$   \\
EN & GPT-4o    & 44.72 & DeepSeek    & 12.29 & $-$33.33 & [$-$38.10, $-$28.57] & $<$0.001 & $<$\textbf{0.001} & $^{***}$ \\
EN & GPT-4o    & 44.72 & Qwen 9B     &  3.20 & $-$40.00 & [$-$42.86, $-$38.10] & $<$0.001 & $<$\textbf{0.001} & $^{***}$ \\
EN & GPT-5.4   &  9.09 & Son.\ 4.6   & 17.68 & +9.52    & [4.76, 14.29]        & $<$0.001 & \textbf{0.001}    & $^{**}$  \\
EN & GPT-5.4   &  9.09 & Gem.\ 3 Fl. & 35.21 & +22.22   & [16.67, 28.57]       & $<$0.001 & $<$\textbf{0.001} & $^{***}$ \\
EN & GPT-5.4   &  9.09 & DeepSeek    & 12.29 & +0.00    & [0.00, 4.76]         & 0.136    & 0.952             &          \\
EN & GPT-5.4   &  9.09 & Qwen 9B     &  3.20 & +0.00    & [$-$4.76, 0.00]      & $<$0.001 & \textbf{0.005}    & $^{**}$  \\
EN & Son.\ 4.6 & 17.68 & Gem.\ 3 Fl. & 35.21 & +14.29   & [9.52, 16.67]        & $<$0.001 & $<$\textbf{0.001} & $^{***}$ \\
EN & Son.\ 4.6 & 17.68 & DeepSeek    & 12.29 & $-$8.33  & [$-$9.52, $-$4.17]   & $<$0.001 & \textbf{0.008}    & $^{**}$  \\
EN & Son.\ 4.6 & 17.68 & Qwen 9B     &  3.20 & $-$14.29 & [$-$16.67, $-$14.29] & $<$0.001 & $<$\textbf{0.001} & $^{***}$ \\
EN & Gem.\ 3 Fl. & 35.21 & DeepSeek  & 12.29 & $-$22.22 & [$-$23.81, $-$16.67] & $<$0.001 & $<$\textbf{0.001} & $^{***}$ \\
EN & Gem.\ 3 Fl. & 35.21 & Qwen 9B   &  3.20 & $-$28.57 & [$-$28.57, $-$23.81] & $<$0.001 & $<$\textbf{0.001} & $^{***}$ \\
EN & DeepSeek  & 12.29 & Qwen 9B     &  3.20 & $-$4.76  & [$-$9.52, $-$4.76]   & $<$0.001 & $<$\textbf{0.001} & $^{***}$ \\
\midrule
%
AR & GPT-4o    & 26.25 & GPT-5.4     & 18.40 & $-$4.76  & [$-$9.52, $-$4.76]   & $<$0.001 & \textbf{0.030}    & $^{*}$   \\
AR & GPT-4o    & 26.25 & Son.\ 4.6   & 14.53 & $-$6.67  & [$-$16.67, $-$3.70]  & $<$0.001 & \textbf{0.003}    & $^{**}$  \\
AR & GPT-4o    & 26.25 & Gem.\ 3 Fl. & 51.72 & +23.81   & [19.05, 28.57]       & $<$0.001 & $<$\textbf{0.001} & $^{***}$ \\
AR & GPT-4o    & 26.25 & DeepSeek    &  6.02 & $-$19.05 & [$-$22.22, $-$16.67] & $<$0.001 & $<$\textbf{0.001} & $^{***}$ \\
AR & GPT-4o    & 26.25 & Qwen 9B     &  4.32 & $-$19.05 & [$-$23.81, $-$16.67] & $<$0.001 & $<$\textbf{0.001} & $^{***}$ \\
AR & GPT-5.4   & 18.40 & Son.\ 4.6   & 14.53 & +0.00    & [$-$4.76, 4.17]      & 0.123    & 0.952             &          \\
AR & GPT-5.4   & 18.40 & Gem.\ 3 Fl. & 51.72 & +33.33   & [27.78, 38.10]       & $<$0.001 & $<$\textbf{0.001} & $^{***}$ \\
AR & GPT-5.4   & 18.40 & DeepSeek    &  6.02 & $-$9.52  & [$-$14.29, $-$9.52]  & $<$0.001 & $<$\textbf{0.001} & $^{***}$ \\
AR & GPT-5.4   & 18.40 & Qwen 9B     &  4.32 & $-$11.11 & [$-$19.05, $-$9.52]  & $<$0.001 & $<$\textbf{0.001} & $^{***}$ \\
AR & Son.\ 4.6 & 14.53 & Gem.\ 3 Fl. & 51.72 & +33.33   & [26.67, 40.00]       & $<$0.001 & $<$\textbf{0.001} & $^{***}$ \\
AR & Son.\ 4.6 & 14.53 & DeepSeek    &  6.02 & $-$9.52  & [$-$14.29, $-$4.76]  & $<$0.001 & $<$\textbf{0.001} & $^{***}$ \\
AR & Son.\ 4.6 & 14.53 & Qwen 9B     &  4.32 & $-$13.33 & [$-$14.29, $-$4.76]  & $<$0.001 & $<$\textbf{0.001} & $^{***}$ \\
AR & Gem.\ 3 Fl. & 51.72 & DeepSeek  &  6.02 & $-$46.67 & [$-$52.38, $-$38.10] & $<$0.001 & $<$\textbf{0.001} & $^{***}$ \\
AR & Gem.\ 3 Fl. & 51.72 & Qwen 9B   &  4.32 & $-$46.67 & [$-$55.56, $-$38.89] & $<$0.001 & $<$\textbf{0.001} & $^{***}$ \\
AR & DeepSeek  &  6.02 & Qwen 9B     &  4.32 & +0.00    & [0.00, 0.00]         & 0.015    & 0.594             &          \\
\midrule
%
TR & GPT-4o    & 40.53 & GPT-5.4     & 20.67 & $-$21.43 & [$-$23.81, $-$16.67] & $<$0.001 & $<$\textbf{0.001} & $^{***}$ \\
TR & GPT-4o    & 40.53 & Son.\ 4.6   & 15.99 & $-$23.81 & [$-$28.57, $-$20.83] & $<$0.001 & $<$\textbf{0.001} & $^{***}$ \\
TR & GPT-4o    & 40.53 & Gem.\ 3 Fl. & 40.30 & $-$0.00  & [$-$4.76, 5.56]      & 0.722    & 0.952             &          \\
TR & GPT-4o    & 40.53 & DeepSeek    & 13.33 & $-$27.78 & [$-$33.33, $-$22.22] & $<$0.001 & $<$\textbf{0.001} & $^{***}$ \\
TR & GPT-4o    & 40.53 & Qwen 9B     &  3.95 & $-$33.33 & [$-$42.86, $-$33.33] & $<$0.001 & $<$\textbf{0.001} & $^{***}$ \\
TR & GPT-5.4   & 20.67 & Son.\ 4.6   & 15.99 & +0.00    & [$-$4.76, 4.76]      & 0.212    & 0.952             &          \\
TR & GPT-5.4   & 20.67 & Gem.\ 3 Fl. & 40.30 & +16.67   & [14.29, 23.81]       & $<$0.001 & $<$\textbf{0.001} & $^{***}$ \\
TR & GPT-5.4   & 20.67 & DeepSeek    & 13.33 & $-$4.76  & [$-$9.52, $-$4.76]   & $<$0.001 & \textbf{0.004}    & $^{**}$  \\
TR & GPT-5.4   & 20.67 & Qwen 9B     &  3.95 & $-$14.29 & [$-$19.05, $-$9.52]  & $<$0.001 & $<$\textbf{0.001} & $^{***}$ \\
TR & Son.\ 4.6 & 15.99 & Gem.\ 3 Fl. & 40.30 & +19.05   & [14.29, 27.78]       & $<$0.001 & $<$\textbf{0.001} & $^{***}$ \\
TR & Son.\ 4.6 & 15.99 & DeepSeek    & 13.33 & $-$4.76  & [$-$9.52, 0.00]      & 0.016    & 0.594             &          \\
TR & Son.\ 4.6 & 15.99 & Qwen 9B     &  3.95 & $-$14.29 & [$-$14.29, $-$11.11] & $<$0.001 & $<$\textbf{0.001} & $^{***}$ \\
TR & Gem.\ 3 Fl. & 40.30 & DeepSeek  & 13.33 & $-$27.78 & [$-$33.33, $-$22.22] & $<$0.001 & $<$\textbf{0.001} & $^{***}$ \\
TR & Gem.\ 3 Fl. & 40.30 & Qwen 9B   &  3.95 & $-$33.33 & [$-$38.10, $-$27.78] & $<$0.001 & $<$\textbf{0.001} & $^{***}$ \\
TR & DeepSeek  & 13.33 & Qwen 9B     &  3.95 & $-$4.76  & [$-$9.52, 0.00]      & $<$0.001 & $<$\textbf{0.001} & $^{***}$ \\
\midrule
%
ZH & GPT-4o    & 41.53 & GPT-5.4     &  9.05 & $-$33.33 & [$-$38.10, $-$23.81] & $<$0.001 & $<$\textbf{0.001} & $^{***}$ \\
ZH & GPT-4o    & 41.53 & Son.\ 4.6   & 13.14 & $-$27.78 & [$-$33.33, $-$20.00] & $<$0.001 & $<$\textbf{0.001} & $^{***}$ \\
ZH & GPT-4o    & 41.53 & Gem.\ 3 Fl. & 36.72 & $-$4.76  & [$-$11.11, 0.00]     & 0.063    & 0.952             &          \\
ZH & GPT-4o    & 41.53 & DeepSeek    & 14.07 & $-$27.78 & [$-$33.33, $-$20.83] & $<$0.001 & $<$\textbf{0.001} & $^{***}$ \\
ZH & GPT-4o    & 41.53 & Qwen 9B     &  3.66 & $-$38.10 & [$-$38.89, $-$33.33] & $<$0.001 & $<$\textbf{0.001} & $^{***}$ \\
ZH & GPT-5.4   &  9.05 & Son.\ 4.6   & 13.14 & +4.17    & [0.00, 11.11]        & 0.023    & 0.837             &          \\
ZH & GPT-5.4   &  9.05 & Gem.\ 3 Fl. & 36.72 & +22.22   & [19.05, 28.57]       & $<$0.001 & $<$\textbf{0.001} & $^{***}$ \\
ZH & GPT-5.4   &  9.05 & DeepSeek    & 14.07 & +4.76    & [0.00, 5.56]         & 0.014    & 0.570             &          \\
ZH & GPT-5.4   &  9.05 & Qwen 9B     &  3.66 & +0.00    & [$-$4.76, 0.00]      & $<$0.001 & \textbf{0.002}    & $^{**}$  \\
ZH & Son.\ 4.6 & 13.14 & Gem.\ 3 Fl. & 36.72 & +20.00   & [14.29, 28.57]       & $<$0.001 & $<$\textbf{0.001} & $^{***}$ \\
ZH & Son.\ 4.6 & 13.14 & DeepSeek    & 14.07 & +0.00    & [$-$4.76, 0.00]      & 0.530    & 0.952             &          \\
ZH & Son.\ 4.6 & 13.14 & Qwen 9B     &  3.66 & $-$9.52  & [$-$14.29, $-$4.76]  & $<$0.001 & $<$\textbf{0.001} & $^{***}$ \\
ZH & Gem.\ 3 Fl. & 36.72 & DeepSeek  & 14.07 & $-$22.22 & [$-$23.81, $-$14.29] & $<$0.001 & $<$\textbf{0.001} & $^{***}$ \\
ZH & Gem.\ 3 Fl. & 36.72 & Qwen 9B   &  3.66 & $-$28.57 & [$-$33.33, $-$23.81] & $<$0.001 & $<$\textbf{0.001} & $^{***}$ \\
ZH & DeepSeek  & 14.07 & Qwen 9B     &  3.66 & $-$5.56  & [$-$9.52, $-$4.76]   & $<$0.001 & $<$\textbf{0.001} & $^{***}$ \\
\midrule
%
HI & GPT-4o    & 44.02 & GPT-5.4     & 23.30 & $-$22.22 & [$-$25.00, $-$16.67] & $<$0.001 & $<$\textbf{0.001} & $^{***}$ \\
HI & GPT-4o    & 44.02 & Son.\ 4.6   & 13.30 & $-$33.33 & [$-$38.10, $-$26.67] & $<$0.001 & $<$\textbf{0.001} & $^{***}$ \\
HI & GPT-4o    & 44.02 & Gem.\ 3 Fl. & 53.22 & +9.52    & [0.00, 14.29]        & 0.009    & 0.359             &          \\
HI & GPT-4o    & 44.02 & DeepSeek    & 17.77 & $-$28.57 & [$-$33.33, $-$23.81] & $<$0.001 & $<$\textbf{0.001} & $^{***}$ \\
HI & GPT-4o    & 44.02 & Qwen 9B     &  4.11 & $-$38.89 & [$-$44.44, $-$33.33] & $<$0.001 & $<$\textbf{0.001} & $^{***}$ \\
HI & GPT-5.4   & 23.30 & Son.\ 4.6   & 13.30 & $-$5.56  & [$-$13.33, $-$4.17]  & $<$0.001 & \textbf{0.001}    & $^{**}$  \\
HI & GPT-5.4   & 23.30 & Gem.\ 3 Fl. & 53.22 & +33.33   & [22.22, 33.33]       & $<$0.001 & $<$\textbf{0.001} & $^{***}$ \\
HI & GPT-5.4   & 23.30 & DeepSeek    & 17.77 & $-$5.56  & [$-$9.52, 0.00]      & 0.003    & 0.128             &          \\
HI & GPT-5.4   & 23.30 & Qwen 9B     &  4.11 & $-$19.05 & [$-$20.00, $-$14.29] & $<$0.001 & $<$\textbf{0.001} & $^{***}$ \\
HI & Son.\ 4.6 & 13.30 & Gem.\ 3 Fl. & 53.22 & +40.00   & [28.57, 44.44]       & $<$0.001 & $<$\textbf{0.001} & $^{***}$ \\
HI & Son.\ 4.6 & 13.30 & DeepSeek    & 17.77 & +4.76    & [0.00, 5.56]         & 0.036    & 0.952             &          \\
HI & Son.\ 4.6 & 13.30 & Qwen 9B     &  4.11 & $-$9.52  & [$-$12.50, $-$4.76]  & $<$0.001 & $<$\textbf{0.001} & $^{***}$ \\
HI & Gem.\ 3 Fl. & 53.22 & DeepSeek  & 17.77 & $-$38.10 & [$-$40.74, $-$28.57] & $<$0.001 & $<$\textbf{0.001} & $^{***}$ \\
HI & Gem.\ 3 Fl. & 53.22 & Qwen 9B   &  4.11 & $-$47.62 & [$-$60.00, $-$38.10] & $<$0.001 & $<$\textbf{0.001} & $^{***}$ \\
HI & DeepSeek  & 17.77 & Qwen 9B     &  4.11 & $-$11.11 & [$-$14.29, $-$9.52]  & $<$0.001 & $<$\textbf{0.001} & $^{***}$ \\
\end{longtable}
\endgroup

\section{Arabic Performance Analysis}
\label{app:arabic_analysis}

Arabic provides the clearest example of backbone-dependent language effects. Under GPT-4o, Arabic is the weakest setting, largely due to stricter evidence weighting: the judge more often rejects placeholder artifacts, unresolved warnings, and trajectory-only claims unless supported by direct file content.

Gemini behaves differently, achieving the highest Arabic satisfaction (51.72\%, $p<0.001$, 95\% CI $[19.05,28.57]$ vs.\ GPT-4o). The advantage persists at higher thresholds: Gemini exceeds 20\%, 50\%, and 70\% satisfaction on 89.09\%, 47.88\%, and 25.45\% of tasks, respectively, and reaches 6.67\% TSR.

The requirement-type breakdown clarifies the gap. In Arabic, Gemini performs strongly across \textit{Data Loading} (70.97\%), \textit{Training} (80.00\%), \textit{Visualization} (63.64\%), \textit{Model Construction} (33.33\%), and \textit{Evaluation Metrics} (38.00\%). In contrast, GPT-4o drops sharply in semantically dense categories such as \textit{Model Construction} (14.06\%) and \textit{Evaluation Metrics} (14.67\%). The Arabic results therefore reflect cross-backbone differences rather than inherent language difficulty.

\section{GPT-4o Baseline: Detailed Multilingual Figures}
\label{app:gpt4o_multilingual_figures}

\begin{figure}[t]
\centering
\includegraphics[width=\linewidth]{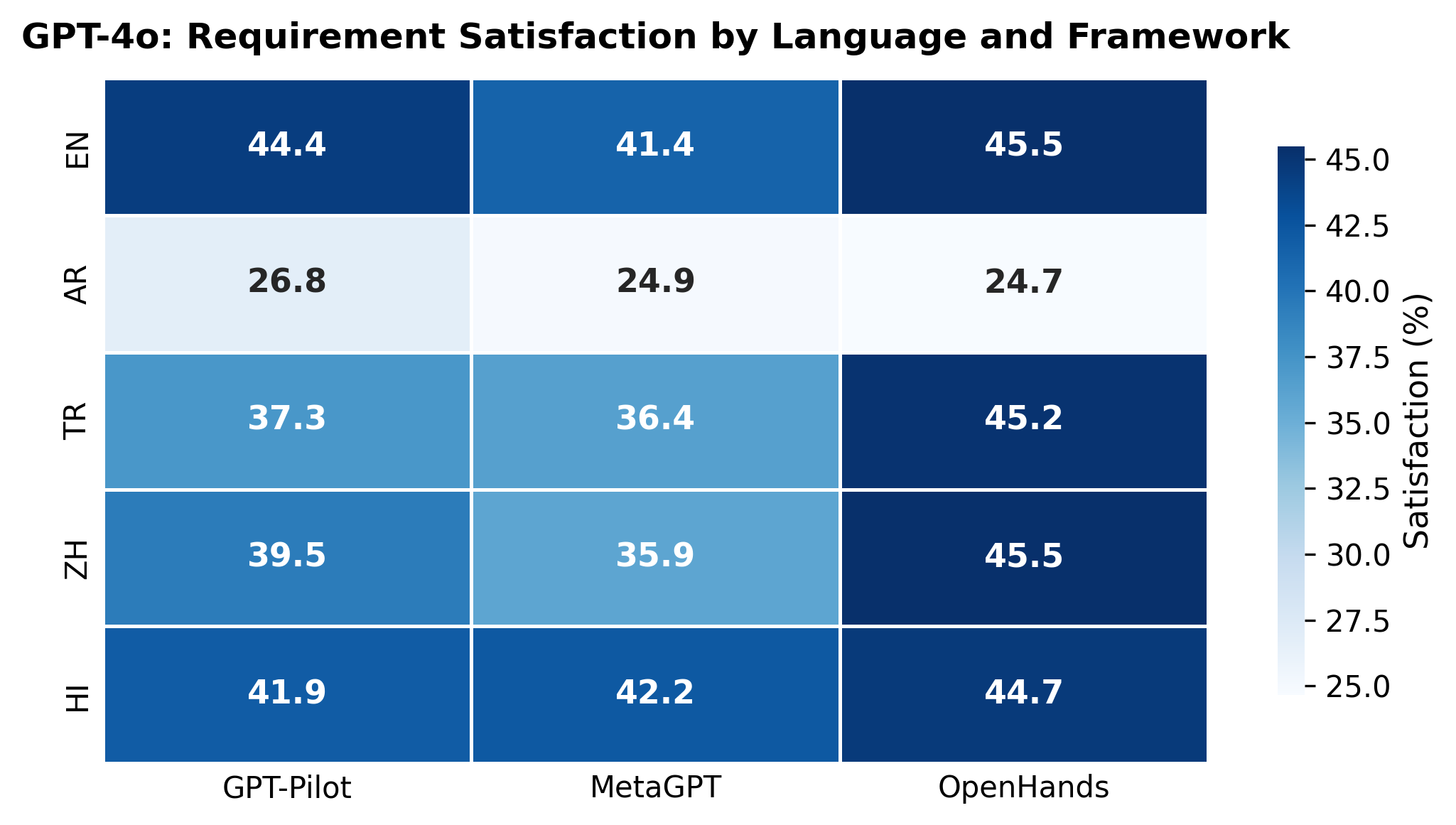}
\caption{Requirement satisfaction (\%) by language and framework for the GPT-4o baseline, shown as a heatmap where darker blue indicates higher satisfaction. Arabic is the weakest language across all three frameworks, while English, Hindi, Turkish, and Chinese form a stronger cluster with smaller cross-framework variation.}
\label{fig:multilingual_satisfaction_heatmap}
\end{figure}

\begin{figure}[t]
\centering
\includegraphics[width=0.6\linewidth]{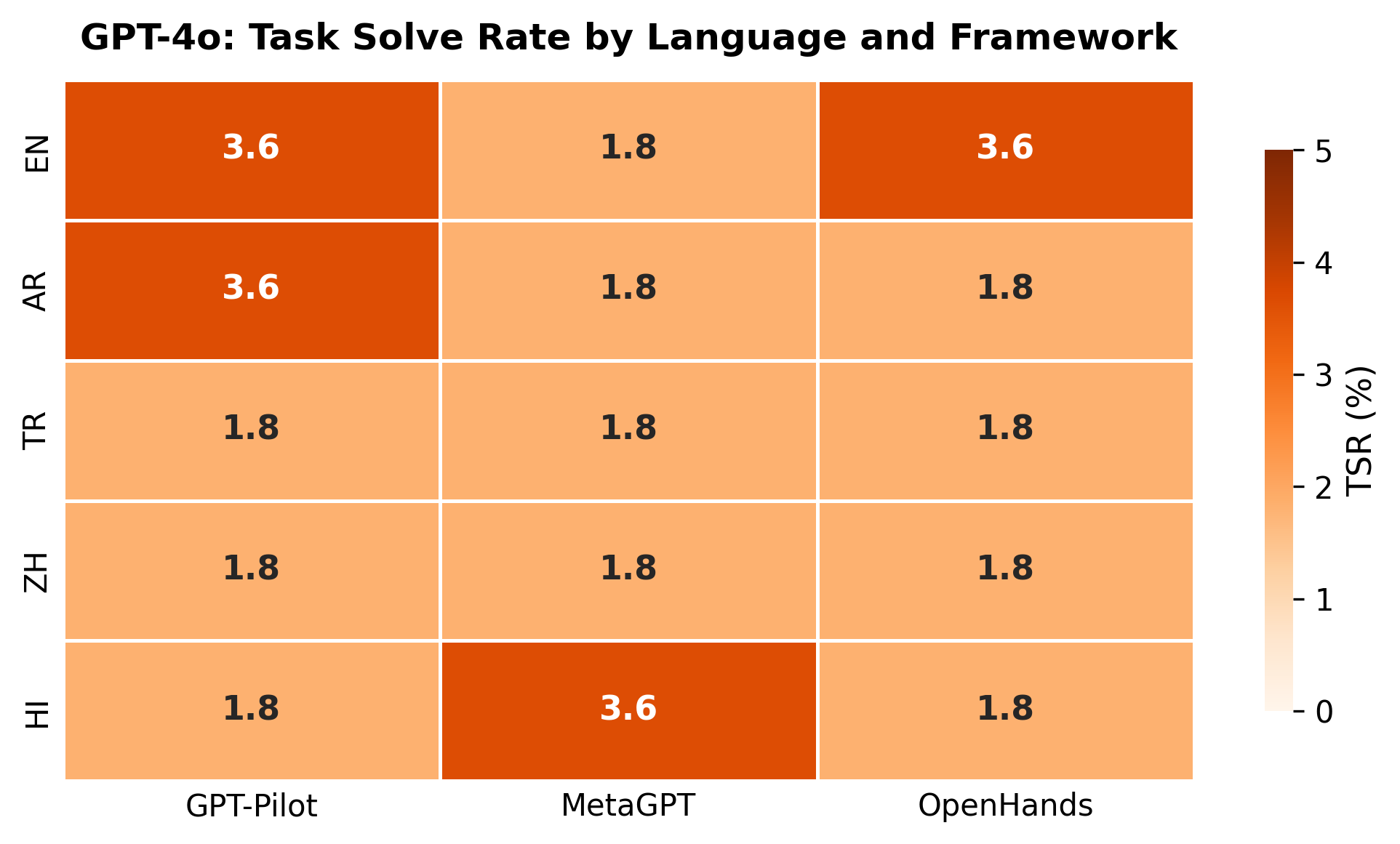}
\caption{Task Solve Rate (\%) by language and framework for the GPT-4o baseline, shown as a heatmap where darker orange indicates higher completion. TSR remains low across all settings (1.82\% to 3.64\%), confirming that full task completion is rare even under the strongest backbone, despite substantial partial progress.}
\label{fig:multilingual_tsr_heatmap}
\end{figure}

\begin{figure}[t]
\centering
\includegraphics[width=\linewidth]{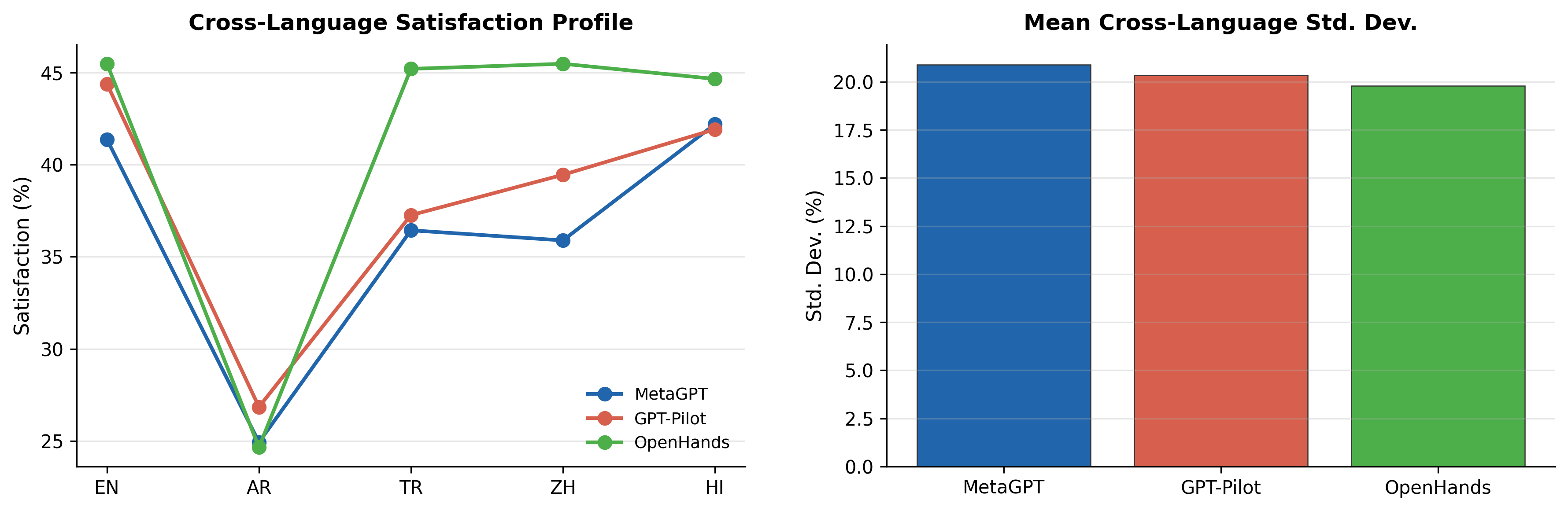}
\caption{Cross-language satisfaction profiles (left) and cross-language variability (right) for the GPT-4o baseline. The left panel shows that all three frameworks exhibit the same Arabic dip, while OpenHands is the most stable across non-Arabic languages. The right panel quantifies cross-language standard deviation per framework.}
\label{fig:multilingual_framework_robustness}
\end{figure}

\begin{figure}[t]
\centering
\includegraphics[width=\linewidth]{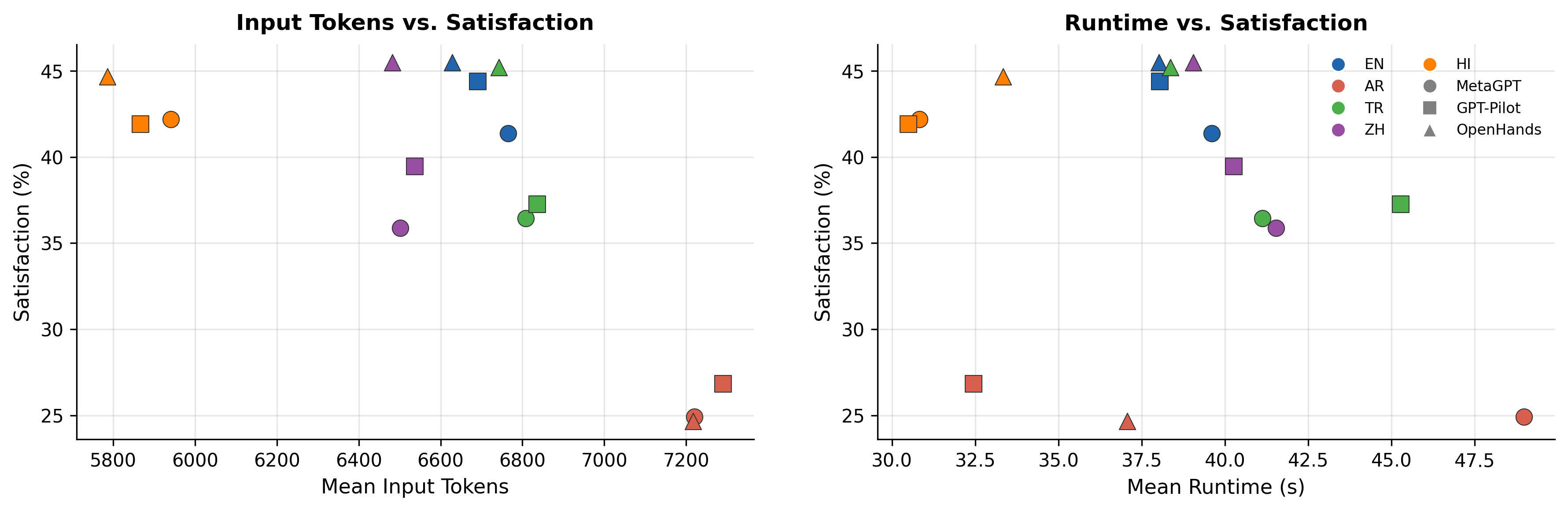}
\caption{Input tokens vs.\ satisfaction (left) and runtime vs.\ satisfaction (right) for the GPT-4o baseline. Each point is a language-framework pair. Arabic uses the most input tokens but achieves the lowest satisfaction, while Hindi uses the fewest tokens and achieves among the highest satisfaction. This visual evidence is consistent with the non-significant Spearman correlations reported in the Discussion.}
\label{fig:multilingual_efficiency_scatter}
\end{figure}

\begin{figure}[t]
\centering
\includegraphics[width=\linewidth]{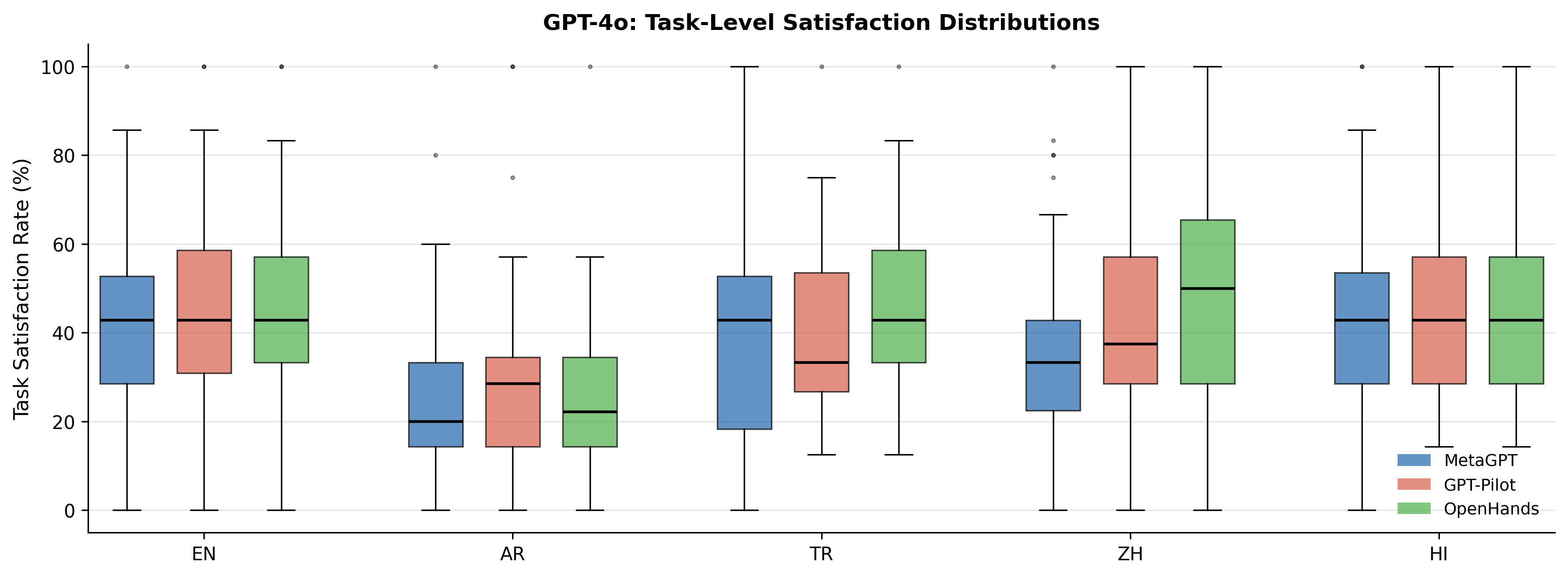}
\caption{Task-level requirement satisfaction distributions by language and framework for the GPT-4o baseline. Box plots show median, interquartile range, and outliers across the 55 tasks. The large spread within each setting confirms that task-level heterogeneity is substantial, reinforcing the importance of reporting variance alongside means.}
\label{fig:multilingual_task_distribution_boxplot}
\end{figure}

\begin{figure}[t]
\centering
\includegraphics[width=\linewidth]{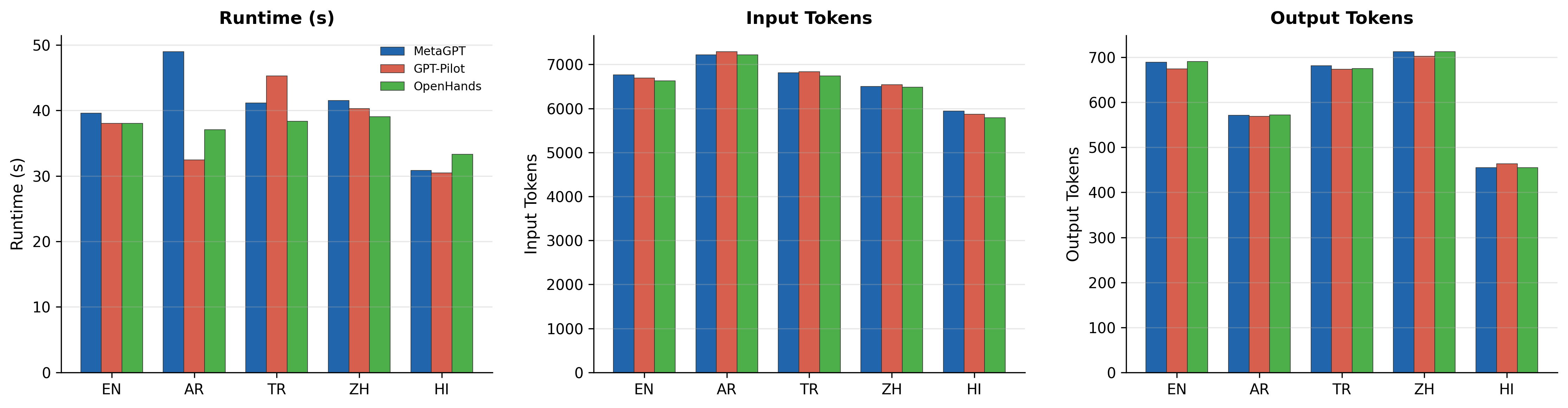}
\caption{Runtime and token usage panels by language and framework for the GPT-4o baseline. These panels complement Table~\ref{tab:backbone_overall} by showing how resource consumption varies across multilingual settings within a single backbone.}
\label{fig:multilingual_runtime_token_panels}
\end{figure}

Figures~\ref{fig:multilingual_satisfaction_heatmap}--\ref{fig:multilingual_runtime_token_panels} provide framework-resolved detail for the GPT-4o baseline, which serves as the primary reference point for the multilingual analysis.

\clearpage
\section{Full Numerical Tables for Main-Text Figures}
\label{app:dense_tables}
We report the full numerical values corresponding to the main-text figures in Tables~\ref{tab:language_backbone_success} and~\ref{tab:requirement_type_sensitivity}.

\begin{table*}[!t]
\centering
\footnotesize
\renewcommand{\arraystretch}{1.3}
\setlength{\tabcolsep}{2.8pt}
\caption{Percentile success rates (\%) by language and backbone, averaged over three frameworks. Each backbone group reports the fraction of tasks exceeding 20\%, 50\%, 70\%, and 100\% satisfaction (100\% = TSR). Numerical values underlying Figure~\ref{fig:percentile_success_main}. \colorbox{gold!30}{\strut Gold} = best per row-threshold; \colorbox{silver!40}{\strut Silver} = second best.}
\label{tab:language_backbone_success}
\resizebox{\textwidth}{!}{%
\begin{tabular}{@{}l rrrr@{\hskip 6pt} rrrr@{\hskip 6pt} rrrr@{\hskip 6pt} rrrr@{\hskip 6pt} rrrr@{\hskip 6pt} rrrr@{}}
\toprule
& \multicolumn{4}{c}{\textbf{GPT-4o}}
& \multicolumn{4}{c}{\textbf{GPT-5.4}}
& \multicolumn{4}{c}{\textbf{Sonnet 4.6}}
& \multicolumn{4}{c}{\textbf{Gemini 3 Flash}}
& \multicolumn{4}{c}{\textbf{DeepSeek V3.2}}
& \multicolumn{4}{c}{\textbf{Qwen3.5-9B}} \\
\cmidrule(lr){2-5}\cmidrule(lr){6-9}\cmidrule(lr){10-13}\cmidrule(lr){14-17}\cmidrule(lr){18-21}\cmidrule(lr){22-25}
\textbf{Lang.}
& {\scriptsize$>$20} & {\scriptsize$>$50} & {\scriptsize$>$70} & {\scriptsize 100}
& {\scriptsize$>$20} & {\scriptsize$>$50} & {\scriptsize$>$70} & {\scriptsize 100}
& {\scriptsize$>$20} & {\scriptsize$>$50} & {\scriptsize$>$70} & {\scriptsize 100}
& {\scriptsize$>$20} & {\scriptsize$>$50} & {\scriptsize$>$70} & {\scriptsize 100}
& {\scriptsize$>$20} & {\scriptsize$>$50} & {\scriptsize$>$70} & {\scriptsize 100}
& {\scriptsize$>$20} & {\scriptsize$>$50} & {\scriptsize$>$70} & {\scriptsize 100} \\
\midrule
EN & \gbest{85.45} & \gbest{33.94} & \gbest{10.91} & \sbest{3.03} & 9.09 & 1.82 & 1.82 & 1.21 & 25.45 & 3.64 & 3.64 & 0.00 & \sbest{62.42} & \sbest{23.64} & \sbest{9.09} & \gbest{4.24} & 16.97 & 4.24 & 2.42 & 0.00 & 4.24 & 1.21 & 0.61 & 0.00 \\
AR & \sbest{50.91} & \sbest{8.48} & \sbest{3.64} & \sbest{2.42} & 35.76 & 4.85 & 2.42 & 0.61 & 21.21 & 3.64 & 3.64 & 0.00 & \gbest{89.09} & \gbest{47.88} & \gbest{25.45} & \gbest{6.67} & 8.48 & 3.03 & 0.61 & 0.00 & 4.85 & 1.82 & 1.82 & 0.00 \\
TR & \gbest{80.61} & \gbest{27.88} & \sbest{8.48} & \sbest{1.82} & 37.58 & 8.48 & 4.85 & 0.61 & 24.24 & 3.64 & 3.64 & 0.00 & \sbest{78.18} & \sbest{25.45} & \gbest{12.12} & \gbest{4.85} & 23.03 & 7.27 & 2.42 & 0.00 & 5.45 & 1.82 & 0.61 & 0.00 \\
ZH & \gbest{80.00} & \gbest{30.30} & \gbest{12.73} & \sbest{1.82} & 12.12 & 2.42 & 1.82 & 0.61 & 14.55 & 3.64 & 3.64 & 0.00 & \sbest{70.30} & \sbest{22.42} & \sbest{10.91} & \gbest{2.42} & 16.97 & 6.67 & 3.64 & 0.00 & 5.45 & 1.21 & 0.61 & 0.00 \\
HI & \gbest{88.48} & \sbest{31.52} & \sbest{10.30} & \sbest{2.42} & 46.67 & 7.88 & 2.42 & 0.00 & 16.36 & 3.64 & 3.64 & 0.00 & \sbest{86.67} & \gbest{50.30} & \gbest{33.33} & \gbest{5.45} & 30.30 & 6.67 & 3.03 & 0.00 & 6.67 & 1.82 & 1.21 & 0.00 \\
\bottomrule
\end{tabular}%
}
\end{table*}

\begin{table*}[!t]
\centering
\footnotesize
\renewcommand{\arraystretch}{1.3}
\setlength{\tabcolsep}{2.0pt}
\caption{Requirement-type sensitivity (\%) across languages and backbones. Requirements grouped into six functional categories; values pooled across three frameworks. Numerical values underlying Figure~\ref{fig:requirement_type_sensitivity}. \colorbox{gold!30}{\strut Gold} = best per row; \colorbox{silver!40}{\strut Silver} = second best.}
\label{tab:requirement_type_sensitivity}
\resizebox{\textwidth}{!}{%
\begin{tabular}{@{}l rrrrr@{\hskip 6pt} rrrrr@{\hskip 6pt} rrrrr@{\hskip 6pt} rrrrr@{\hskip 6pt} rrrrr@{\hskip 6pt} rrrrr@{}}
\toprule
& \multicolumn{5}{c}{\textbf{GPT-4o}}
& \multicolumn{5}{c}{\textbf{GPT-5.4}}
& \multicolumn{5}{c}{\textbf{Sonnet 4.6}}
& \multicolumn{5}{c}{\textbf{Gemini 3 Flash}}
& \multicolumn{5}{c}{\textbf{DeepSeek V3.2}}
& \multicolumn{5}{c}{\textbf{Qwen3.5-9B}} \\
\cmidrule(lr){2-6}\cmidrule(lr){7-11}\cmidrule(lr){12-16}\cmidrule(lr){17-21}\cmidrule(lr){22-26}\cmidrule(lr){27-31}
\textbf{Req.\ Type (\#)}
& {\scriptsize EN} & {\scriptsize AR} & {\scriptsize TR} & {\scriptsize ZH} & {\scriptsize HI}
& {\scriptsize EN} & {\scriptsize AR} & {\scriptsize TR} & {\scriptsize ZH} & {\scriptsize HI}
& {\scriptsize EN} & {\scriptsize AR} & {\scriptsize TR} & {\scriptsize ZH} & {\scriptsize HI}
& {\scriptsize EN} & {\scriptsize AR} & {\scriptsize TR} & {\scriptsize ZH} & {\scriptsize HI}
& {\scriptsize EN} & {\scriptsize AR} & {\scriptsize TR} & {\scriptsize ZH} & {\scriptsize HI}
& {\scriptsize EN} & {\scriptsize AR} & {\scriptsize TR} & {\scriptsize ZH} & {\scriptsize HI} \\
\midrule
Data Load.\ (62) & 46.24 & \sbest{66.13} & 59.68 & 50.54 & 60.22 & 3.76 & 5.91 & 13.44 & 4.30 & 16.67 & 2.69 & 3.23 & 3.23 & 3.23 & 3.23 & 25.81 & \gbest{70.97} & 38.17 & 31.18 & 61.83 & 5.38 & 1.61 & 6.99 & 8.06 & 12.37 & 1.08 & 2.15 & 1.61 & 1.61 & 1.61 \\
Preproc.\ (70) & 39.05 & 17.14 & 18.10 & \gbest{45.71} & \sbest{41.90} & 4.29 & 6.19 & 9.05 & 2.86 & 8.57 & 7.14 & 5.24 & 6.19 & 4.29 & 5.71 & 18.57 & 33.33 & 24.76 & 15.24 & 39.05 & 4.76 & 2.86 & 5.24 & 7.14 & 6.19 & 1.90 & 2.86 & 2.86 & 1.90 & 2.86 \\
Model Constr.\ (64) & 29.69 & 14.06 & \sbest{38.02} & 28.65 & 30.21 & 3.12 & 10.94 & 18.75 & 4.69 & 15.62 & 6.25 & 6.25 & 6.25 & 5.73 & 5.73 & 22.40 & 33.33 & 29.69 & 26.56 & \gbest{39.06} & 10.42 & 4.69 & 11.98 & 10.42 & 15.10 & 2.60 & 3.12 & 4.69 & 2.60 & 4.69 \\
Training (20) & 78.33 & 41.67 & 56.67 & 75.00 & 65.00 & 38.33 & 56.67 & 41.67 & 46.67 & 51.67 & 30.00 & 25.00 & 21.67 & 21.67 & 21.67 & 66.67 & \sbest{80.00} & 71.67 & 73.33 & \gbest{81.67} & 28.33 & 26.67 & 26.67 & 30.00 & 33.33 & 20.00 & 25.00 & 18.33 & 23.33 & 20.00 \\
Eval.\ Metrics (50) & \gbest{58.67} & 14.67 & 38.67 & 33.33 & \sbest{41.33} & 19.33 & 24.00 & 18.00 & 16.67 & 20.67 & 8.00 & 3.33 & 4.00 & 2.00 & 2.00 & 24.00 & 38.00 & 21.33 & 32.00 & 30.00 & 8.67 & 4.00 & 3.33 & 6.00 & 3.33 & 2.00 & 2.67 & 2.00 & 2.00 & 2.00 \\
Visualiz.\ (99) & 40.07 & 15.49 & 40.40 & 34.01 & 37.37 & 7.41 & 25.59 & 27.27 & 6.06 & 36.03 & 41.75 & 35.35 & 40.07 & 31.99 & 31.65 & 56.23 & \sbest{63.64} & 57.91 & 51.85 & \gbest{69.70} & 17.85 & 6.73 & 21.21 & 22.22 & 31.31 & 1.68 & 2.36 & 1.68 & 2.02 & 2.02 \\
\bottomrule
\end{tabular}%
}
\end{table*}

\clearpage
\section{Supplementary Material Note}
Detailed requirement-level tables, task-level runtime statistics, and the full comprehensive multilingual experiment tables are provided in the supplementary material.

\end{document}